\documentclass[10pt,twocolumn,letterpaper]{article}

\usepackage{cvpr}              

\usepackage{cuted}

\usepackage{animate}

\usepackage{subcaption}

\usepackage{xspace}
\usepackage{colortbl}
\usepackage{tabularx,booktabs}
\usepackage{bm}

\usepackage{listings} 

\usepackage{multicol}

\definecolor{cvprblue}{rgb}{0.21,0.49,0.74}
\usepackage[pagebackref,breaklinks,colorlinks,citecolor=cvprblue]{hyperref}

\usepackage[dvipsnames]{xcolor}

\def\paperID{13}
\def\confName{CVPR}
\def\confYear{2024}

\newcommand{\methodshort}{GPT4Motion}

\title{GPT4Motion: Scripting Physical Motions in Text-to-Video Generation via
Blender-Oriented GPT Planning}

\author{Jiaxi Lv$^{1, 2*}$ \quad Yi Huang$^{1, 2*}$ \quad Mingfu Yan$^{1, 2*}$\quad Jiancheng Huang$^{1, 2}$\quad Jianzhuang Liu$^{1}$ 
	\\ 
	Yifan Liu$^{1}$\quad Yafei Wen$^{3}$\quad Xiaoxin Chen$^{3}$\quad Shifeng Chen$^{1\dagger}$
	\\ 
	$^1$Shenzhen Institute of Advanced Technology, Chinese Academy of Sciences, \\ $^2$University of Chinese Academy of Sciences, $^3$VIVO AI Lab  
}

\begin{document}

\def\sizefive{0.18}
\def\jianxi{2mm}

\twocolumn[{%
\maketitle
\renewcommand\twocolumn[1][]{#1}%
    \def\sizefive{0.20}
\vspace{-5mm}
\setlength{\tabcolsep}{0.5pt}
\renewcommand{\arraystretch}{1.15}
\begin{tabular}{c c c c c}
	\multicolumn{5}{c}{\includegraphics[width=0.7\columnwidth]{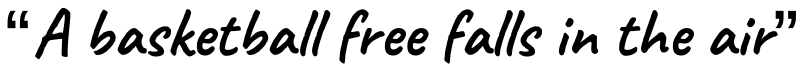}}\\
	\animategraphics[width=\sizefive\textwidth, autoplay, loop]{24}{video_imgs/formal_experiment/basketball/video-A-basketball-fell-out-of-the-air_crop_compress_35/}{0}{100} &
	\animategraphics[width=\sizefive\textwidth, autoplay, loop]{7}{video_imgs/animatediff/}{0}{15} &
	\animategraphics[width=\sizefive\textwidth, autoplay, loop]{7}{video_imgs/Modelscope/}{0}{15} &
	\animategraphics[width=\sizefive\textwidth, autoplay, loop]{3}{video_imgs/tv20/}{0}{7} &
	\animategraphics[width=\sizefive\textwidth, autoplay, loop]{3}{video_imgs/DirecT2V/}{0}{7}\\
	\multicolumn{1}{c}{GPT4Motion}&AnimateDiff \cite{guo2023animatediff} &ModelScope \cite{wang2023modelscope} &Text2Video-Zero \cite{khachatryan2023text2video}&DirecT2V \cite{hong2023large}  \\
\end{tabular}
\vspace{-0.2cm}
\captionof{figure}{Comparison of the video results generated by different text-to-video models with the prompt \textit{``A basketball free falls in the air"}.  {\emph{Best viewed with \href{https://www.adobe.com/acrobat/pdf-reader.html}{Acrobat Reader} for animation. 
}}}
\vspace{3mm}
\label{sec1:comparison_experiment}
}]

\begingroup
\renewcommand{\thefootnote}{}
\footnotetext{$^{}$Work done during the internship of Lv at VIVO AI Lab.}
\footnotetext{$^{*}$Equal contributions.}
\footnotetext{$^{\dagger}$Corresponding author: shifeng.chen@siat.ac.cn}
\endgroup

\begin{abstract}

Recent advances in text-to-video generation have harnessed the power of diffusion models to create visually compelling content conditioned on text prompts. However, they usually encounter high computational costs and often struggle to produce videos with coherent physical motions. To tackle these issues, we propose \methodshort{}, a training-free framework that leverages the planning capability of large language models such as GPT, the physical simulation strength of Blender, and the excellent image generation ability of text-to-image diffusion models to enhance the quality of video synthesis. Specifically, \methodshort{} employs GPT-4 to generate a Blender script based on a user textual prompt, which commands Blender's built-in physics engine to craft fundamental scene components that encapsulate coherent physical motions across frames. Then these components are inputted into Stable Diffusion to generate a video aligned with the textual prompt.
Experimental results on three basic physical motion scenarios, including rigid object drop and collision, cloth draping and swinging, and liquid flow, demonstrate that \methodshort{} can generate high-quality videos efficiently in maintaining motion coherency and entity consistency. \methodshort{} offers new insights in text-to-video research, enhancing its quality and broadening its horizon for future explorations. Our homepage website is \url{https://GPT4Motion.github.io}.
\end{abstract}

\section{Introduction}
\label{sec:intro}

In recent years, the computer vision community has shown increasing interest in generative AI. The rise of diffusion models \cite{sohl2015deep,ho2020denoising, song2021denoising,song2021score} has led to significant advancements in high-quality image generation from textual prompts, commonly known as text-to-image (T2I) synthesis \cite{ramesh2022hierarchical,rombach2022high,saharia2022photorealistic,dhariwal2021diffusion}. Building upon this success, researchers have explored the extension of T2I diffusion models to the realm of text-to-video (T2V) generation and editing. Earlier efforts primarily focus on directly training T2V diffusion models in pixel \cite{ho2022video, singer2023make, ho2022imagen, ge2023preserve} or latent spaces \cite{blattmann2023align, zhou2022magicvideo, yin2023nuwa, esser2023structure, an2023latent, wang2023modelscope, li2023videogen, he2022latent, wang2023lavie}. While such approaches yield promising results, their reliance on extensive datasets \cite{bain2021frozen, xue2022advancing, wang2023internvid} for training leads to heavy computational costs. 
In search of more cost-effective video generation methods, some researchers have proposed mechanisms that adapt existing T2I models for the video domain. For example, Tune-A-Video \cite{wu2023tune} considerably reduces the training effort by fine-tuning a pretrained T2I model like Stable Diffusion \cite{rombach2022high} for video editing. However, it still requires an optimization process for each video generation.

Recent research has shifted towards developing training-free T2V approaches \cite{khachatryan2023text2video, huang2023free} to alleviate the computational burden. For instance, Text2Video-Zero \cite{khachatryan2023text2video} utilizes the pretrained T2I model, Stable Diffusion, to synthesize videos without additional training. While these training-free methods have advanced in reducing resource requirements, they encounter challenges in achieving coherent motions, particularly when using a single user prompt to guide all frames' generation. This limitation can result in videos that lack the continuity of action or miss essential motion details due to the model's limited understanding of the temporal dynamics from a simple abstract description. To address these shortcomings, recent studies \cite{huang2023free, hong2023large} have harnessed the descriptive power of large language models (LLMs) \cite{ouyang2022training, wei2021finetuned}, such as GPT-4 \cite{openai2023gpt} and PaLM2 \cite{anil2023palm}, to generate frame-by-frame descriptions from a single user prompt, aiming to enrich the narrative across the video sequence. Building upon this foundation, subsequent research \cite{lian2023llm, lin2023videodirectorgpt} has taken a step further by instructing LLMs to generate not only detailed descriptions but also explicit spatiotemporal layouts from a single prompt, which then serve as conditions of the T2I diffusion models to generate videos frame by frame. Although the complemented prompts or dynamic layouts improve the video quality over methods relying on a single prompt, it is substantially challenging to ensure motion coherence particularly when there are large motion shifts. 

Motivated by these LLM-assisted methods \cite{feng2023layoutgpt, phung2023grounded, lian2023llm, lin2023videodirectorgpt, huang2023free, hong2023large}, this paper offers a new perspective to handle the problem of motion incoherence. Specifically, we propose \methodshort{}, a training-free framework that leverages the strategic planning capability of GPT-4, the physical simulation strength of Blender\footnote{Blender is a popular open-source 3D creation suite that offers a comprehensive set of tools for 3D modeling, animation, and rendering. See https://www.blender.org/ for details.}, and the excellent image generation ability of Stable Diffusion to enhance the quality of video synthesis. Given a user textual prompt, \methodshort{} begins by deploying GPT-4 to produce Blender scripts that drive the creation of essential video scene elements, including edges and depth maps. Subsequently, these elements are then employed as conditions for Stable Diffusion to generate the final video. This methodology ensures that the resulting video not only faithfully aligns with the textual prompt but also exhibits consistent physical behaviors across all frames, as shown in Figure~\ref{sec1:comparison_experiment}. The contributions of our work are summarized in the following.

\begin{itemize}
	\item We demonstrate the powerful planning capability of GPT-4 in driving Blender to accurately simulate basic physical motion scenes, showing the potential of LLMs to contribute to physics-based video generation tasks.
	
	\item We propose \methodshort{}, a training-free framework that employs scripts generated by GPT-4 for Blender's scene simulation, enabling the generation of temporally coherent videos using the pretrained T2I Stable Diffusion.
 
	\item Experimental results on three basic physical motion scenarios demonstrate that \methodshort{} can efficiently generate high-quality videos which maintain motion coherency and entity consistency.
	
\end{itemize}
\section{Related Work}
\label{sec:related_work}

\subsection{Text-to-Video Generation}
Text-to-video (T2V) generation targets at the creation of videos from textual descriptions. Although significant progress has been made in text-to-image (T2I) synthesis \cite{rombach2022high, dhariwal2021diffusion, ramesh2022hierarchical,2023dalle3,gu2022vector,nichol2021glide,saharia2022photorealistic}, T2V techniques are still in the early stage.
With the advance of diffusion models \cite{ho2020denoising, song2021denoising}, T2V research has shifted towards diffusion-based techniques, which are broadly categorized into training-based \cite{ho2022video, singer2023make, ho2022imagen, ge2023preserve, blattmann2023align, zhou2022magicvideo, yin2023nuwa, esser2023structure, an2023latent, wang2023modelscope, zhang2023show} and training-free \cite{khachatryan2023text2video, huang2023free} approaches. The Video Diffusion Model (VDM) \cite{ho2022video} emerges as a pioneer, adapting the image diffusion U-Net \cite{ronneberger2015u} architecture to a 3D U-Net for joint image and video training. Make-A-Video \cite{singer2023make}, on the other hand, introduces a novel paradigm that learns visual-textual correlations from image-text pairs and acquires motion understanding from unlabelled video data, reducing the need for paired video-text training datasets.

While these methods depend on extensive datasets \cite{bain2021frozen, xue2022advancing, wang2023internvid} for training, recent research \cite{huang2023free, khachatryan2023text2video} focuses on training-free T2V to reduce training costs. For instance, Text2Video-Zero \cite{khachatryan2023text2video} leverages the pretrained Stable Diffusion \cite{rombach2022high} for video synthesis. To maintain consistency across frames, it performs cross-attention between each frame and the first frame. DiffSynth \cite{duan2023diffsynth} proposes a latent in-iteration deflickering framework and a video deflickering algorithm to mitigate flickering and generate coherent videos. Despite their advancements, many T2V models still face challenges such as motion incoherence and entity inconsistency. Addressing these issues, our work introduces a novel approach that integrates the planning power of LLMs with the simulation capability of Blender for T2V synthesis.

\subsection{LLM-Assisted Visual Generations}

Large language models like GPT-4 \cite{openai2023gpt}, PaLM \cite{anil2023palm}, and BLOOM \cite{scao2022bloom} excel in various multimodal tasks\cite{koizumi2020audio,li2022blip,saharia2022photorealistic}.
In the field of text-to-image generation, LLMs have been successfully used to generate prompts \cite{brooks2023instructpix2pix,hao2022optimizing} or to create spatial bounding boxes from textual prompts to control image generation \cite{lian2023llm_image, feng2023layoutgpt, phung2023grounded}. Inspired by these developments, recent efforts \cite{hong2023large, huang2023free, lin2023videodirectorgpt} have started to incorporate LLMs into the T2V realm. For instance, Free-bloom \cite{huang2023free} leverages LLMs to generate detailed frame-by-frame descriptions from a single prompt, thereby enriching the video's narrative. Similarly, LVD \cite{lian2023llm} expands this idea by not only producing detailed descriptions but also creating comprehensive spatiotemporal layouts that guide T2I diffusion models in the frame-by-frame video generation process. Different from them, this paper instructs GPT-4 to generate scripts for Blender to generate scene components which further serve as conditions of Stable Diffusion to synthesize videos.

\subsection{Blender in Deep Learning}
Blender, an open-source 3D creation suite, offers a comprehensive set of tools for 3D modeling, animation, and rendering, enabling the creation of complex and realistic 3D scenes. Beyond its conventional role in visual effects, Blender has also played a key role in deep learning, particularly for generating synthetic data crucial for model training in scenarios where real-world data is lacking, as illustrated by the S2RDA benchmark \cite{tang2023new} for image classification. Additionally, 3D-GPT \cite{sun20233d} instructs LLMs to drive Infinigen \cite{raistrick2023infinite1}, a Python-Blender-based library of generation functions, for procedural 3D modeling. Despite the success achieved in these works, the application of Blender on T2V has not yet been explored. Traditional video creation utilizing Blender often requires much professional technical knowledge and involves complex manual procedures such as texturing, rigging, animation, lighting and compositing.  

Our \methodshort{} simplifies this process by introducing an innovative framework that employs GPT-4 to generate Blender's script, which bypasses the need for manual interaction and intricate scene setup. This not only simplifies the creation process but also ensures the temporal coherence and textual alignment of the generated videos. Through the integration of GPT-4-driven scripting with Blender's advanced simulation capability, \methodshort{} marks substantial progress in the T2V domain, offering a user-friendly and efficient approach to producing high-quality videos.

\begin{figure*}[ht]
    \centering
    \includegraphics[width=\textwidth]{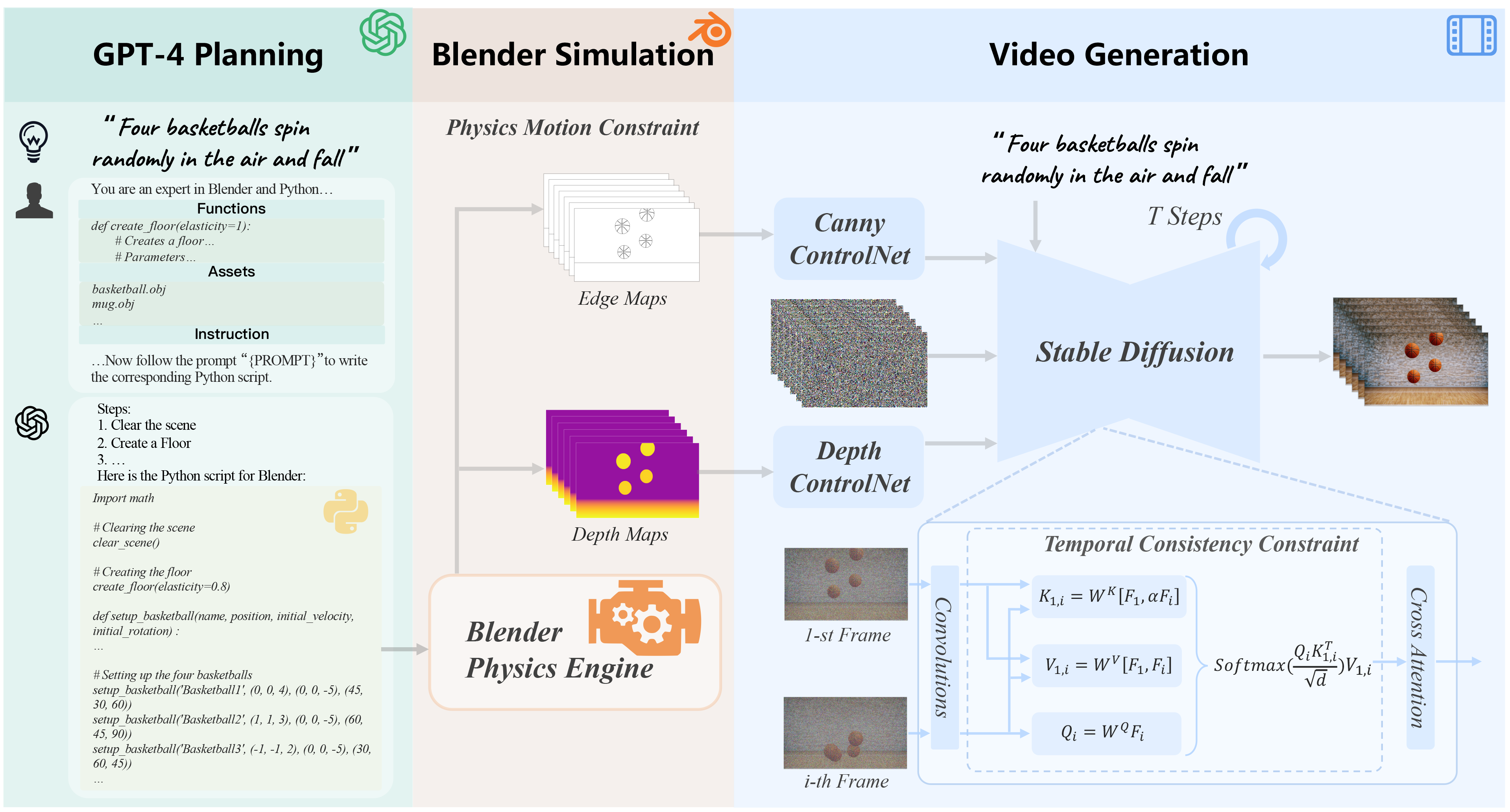}
    \caption{The architecture of our \methodshort{}. First, the user prompt is inserted into our designed prompt template. Then, the Python script generated by GPT-4 drives the Blender physics engine to simulate the corresponding motion, producing sequences of edge maps and depth maps. Finally, two ControlNets are employed to constrain the physical motion of video frames generated by Stable Diffusion, where a temporal consistency constraint is designed to enforce the coherence among frames.}
    \label{framework}
\end{figure*} 

\begin{figure}[ht]
    \centering
    \includegraphics[width=0.49\textwidth]{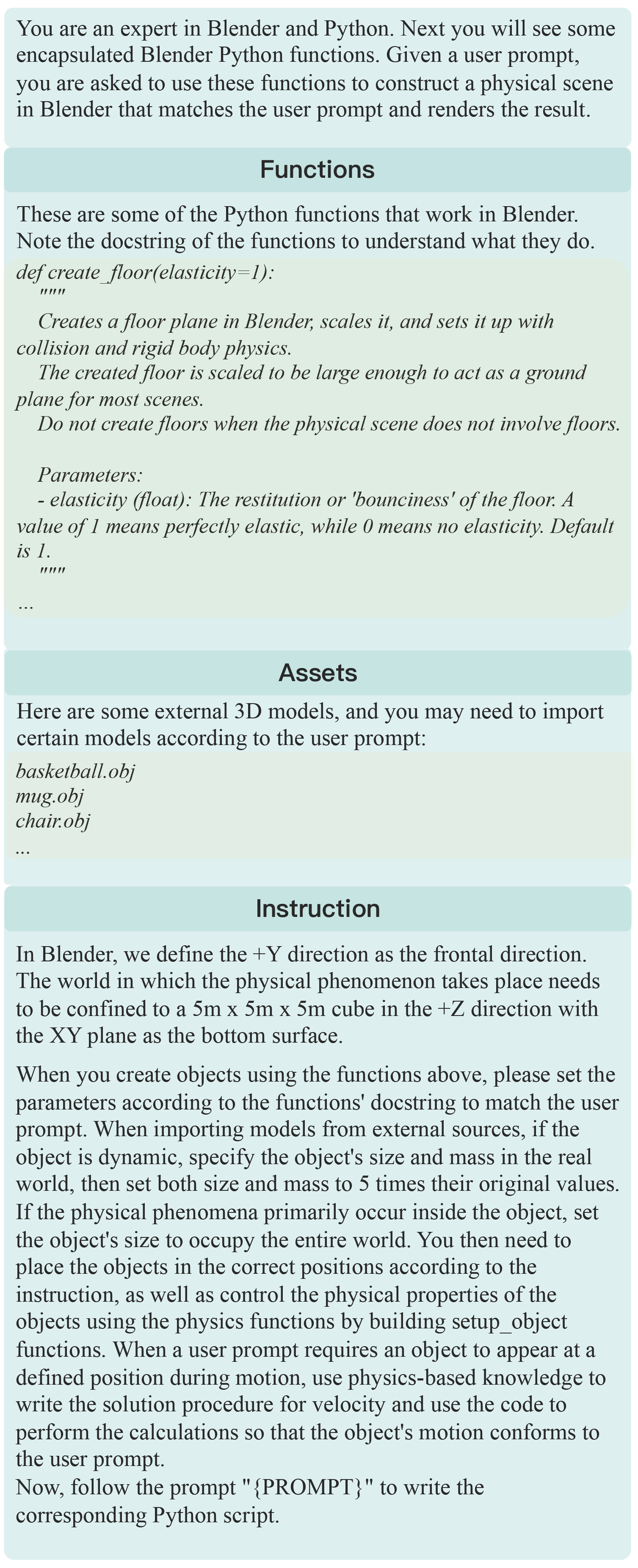}
    \caption{Our prompt template designed for GPT-4. It contains information about functions, external assets, and instruction. The user prompt is inserted into the placeholder ``\{PROMPT\}".}
    \vspace{-1.0cm}
    \label{prompt_gpt4}
\end{figure} 

\section{Method}

\subsection{Task Formulation}
Given a user prompt about some basic physical motion scenario, we aim to generate a physically accurate video. Physical phenomena are often associated with the material of the object. We focus on simulating three common types of object materials encountered in daily life: 1) \textit{Rigid Objects}, such as balls, which maintain their shapes when subjected to forces; 2) \textit{Cloth}, such as flags, characterized by their softness and propensity to flutter; 3) \textit{Liquid}, such as water, which exhibits continuous and deformable motions. Moreover, we give particular attention to several typical motion modes for these materials, including \textit{collisions} (direct impacts between objects), \textit{wind effects} (motion induced by air currents), and \textit{flow} (continuously and easily move in one direction). Simulating these physical scenarios typically involves knowledge of Classical Mechanics \cite{goldstein2002classical}, Fluid Mechanics 
 \cite{kundu2015fluid} and other physical knowledge. Current text-to-video diffusion models struggle to capture this complex physical knowledge through training, thereby failing to produce videos that adhere to physical principles.

To address these challenges, we propose a novel training-free text-to-video generation framework, named \methodshort{}, which is illustrated in Figure \ref{framework}. 
The advantage of our approach is that GPT-4's semantic understanding and code generation capabilities are leveraged to translate the user prompt into a Blender Python script. This script can drive Blender's built-in physics engine to simulate the corresponding physical scene.
We then introduce ControlNet~\cite{zhang2023adding}, which takes as input the dynamic results of the Blender simulation and directs Stable Diffusion to generate each frame of the video. This framework ensures that the generated video is not only consistent with the user prompt, but also physically correct. In the next sections, we describe the details of our framework.

\subsection{Blender Simulations via GPT-4}

GPT-4 is a large language model pre-trained on huge amounts of Internet data with great capability for semantic understanding and code generation. 
We have observed that while GPT-4 has a certain knowledge about the Blender Python API, it still struggles with generating Blender Python scripts based on user prompts. On the one hand, asking GPT-4 to create even a simple 3D model (like a basketball) directly in Blender seems to be an overwhelming task \cite{sun20233d}. On the other hand, because the Blender Python API has fewer resources and its API version is updated quickly, GPT-4 can easily misuse certain functions or make errors due to version differences. To address these issues, we propose the following schemes:
\vspace{-0.3cm}
\paragraph{Leveraging External 3D Models.} Creating 3D models typically requires professional artists to manually craft them, spending substantial time sculpting details, painting fine texture maps, and optimizing the model topology, which GPT-4 cannot independently accomplish. Fortunately, there is a large amount of 3D models available on the Internet\footnote{https://www.blenderkit.com/}. Hence, we have collected common 3D objects from everyday life and can automatically load the 3D models via scripts corresponding to textual prompts.
\vspace{-0.3cm}
\paragraph{Encapsulating Blender Functions.} Although GPT-4 possesses the necessary knowledge of the Blender Python API, writing a lengthy script to render an entire scene remains challenging. We note that for our target scenarios, Blender Python scripts typically consist of several fixed steps, including scene initialization, rendering, object creation and import, and physical effects. Thus, we guide GPT-4 to encapsulate these reusable functions (see the supplement material). By doing so, we have greatly simplified the entire process from user prompts to rendering corresponding physical scenarios. These encapsulated functions can be broadly categorized into three types:

\begin{itemize}
    \item \textit{Scene initialization and rendering functions.} These functions are responsible for clearing the default initial scene and performing the rendering. In Blender, one can set up the simultaneous image outputs of depth, normal, edge, and segmentation for a video. We find that using edge and depth images yields good performance in our framework, so we render these edge and depth images for video generation.
    
    \item \textit{Object creation and import functions.} These functions offer the capability to create basic objects (such as viewpoints, floors, cubes, spheres, etc.) within a Blender scene. In addition to creating simple objects, we also provide import functions that allow users to bring external 3D models into Blender.

    \item \textit{Physics effect functions.} These functions encapsulate the basic physics and material effect settings within Blender. For instance, they can assign different physical types (such as rigid, cloth, or liquid) to objects or set up wind force effects.
\end{itemize}
\vspace{-0.3cm}
\paragraph{Translating User Prompts into Physics.} Figure \ref{prompt_gpt4} shows the general prompt template we design for GPT-4. It includes encapsulated Blender functions, external assets, and instruction. We define the dimensions of the virtual world in the template and provide information about the camera's position and viewpoint. Such information aids GPT-4 in better understanding the layout of the 3D space. Ultimately, the user prompt becomes part of the instruction, directly guiding GPT-4 to generate the corresponding Blender Python script. Finally, with this script, Blender renders the edge and depth image sequences.

\subsection{Video Synthesis with Physical Conditions}
Our goal is to generate a consistent and realistic video based on the user prompt and corresponding physical motion conditions provided by Blender. 
We adopt Stable Diffusion XL (SDXL) \cite{podell2023sdxl}, an upgraded version of Stable Diffusion \cite{rombach2022high}. We made the following modifications to SDXL.
\vspace{-0.3cm}
\paragraph{Physics Motion Constraints.} ControlNet \cite{zhang2023adding} is a network architecture that can control the image generation of a pretrained text-to-image diffusion model with additional conditions, such as edge or depth. However, a single ControlNet is limited to one type of condition. The generation of some physical motion videos requires the control of multiple conditions. For example, when generating a video of a basketball in free fall, its edges can accurately reflect its texture changes, but the edges cannot reflect 3D layout of the scene, resulting in the lack of realism in the video. On the other hand, the depth map of the scene helps address this problem but is unable to capture the texture changes of the basketball. Therefore, we leverage a combination of Canny-edge-based ControlNet and depth-based ControlNet to precisely control the generation of the video. Specifically, we add the intermediate results of the two ControlNets together to serve as the final conditions for SDXL.
\vspace{-0.3cm}
\paragraph{Temporal Consistency Constraint.} To ensure temporal consistency across different frames of a video, we modify the self attention (SA) in the U-Net of SDXL into cross-frame attention (CFA). Specifically, the self attention in the U-Net uses linear projections $W^Q$, $W^K$, and $W^V$ to project the feature $F_i$ of the $i$-th frame (for simplicity, we ignore the time-step $t$) into $Q_i=W^Q F_i$, $K_i=W^K F_i$, and $V_i=W^V F_i$, and perform the self attention calculation:
\begin{equation}
SA(Q_i,K_i,V_i) = \mathrm{Softmax}(Q_i K_i^T / \sqrt{d})V_i,
\end{equation}
where $d$ is a scaling factor. To obtain the cross-frame attention, we concatenate the feature of the frame $F_i$, $i\neq 1$, with the first frame $F_1$ for $K$ and $V$, while keeping $Q$ unchanged:
\begin{equation}
Q_i = W^Q F_i, \; K_{i,1} = W^K [F_1, \alpha F_i], \; V_{i,1} = W^V [F_1, F_i],
\label{sec3:projection}
\end{equation}
and the cross-frame attention operation is:
\begin{equation}
CFA(Q_i,K_{i,1},V_{i,1}) = \mathrm{Softmax}(Q_iK_{i,1}^T / \sqrt{d})V_{i,1},
\label{sec3:cfa}
\end{equation}
where $[\cdot, \cdot]$ denotes the concatenation, and $\alpha \in [0,1]$ is a hyperparameter. We find that increasing $\alpha$ improves the fidelity of the moving object but at the same time brings more flickering; on the contrary, decreasing $\alpha$ reduces the flickering but also decreases the fidelity of the moving object. The cross-frame attention has the effect that the $i$-th frame pays attention to not only itself but also the first frame. Surprisingly, by this cross-frame attention design, the generated video frames exhibit remarkable content consistency.
Additionally, we employ the same initial noise for SDXL to generate all the frames of the video, which further enhances the temporal consistency.
\section{Experiments}

\def\sizethree{0.15}
\def\maxzhen{80}
\def\jianxi{1mm}
\def\sizetwo{0.15}

\def\sizefour{0.11}
\def\jianxifour{0.05mm}

\subsection{Implementation Details}

In our experiments, we use the Stable Diffusion XL 1.0-base model\footnote{\url{https://huggingface.co/stabilityai/stable-diffusion-xl-base-1.0}}, along with Canny-edge-based ControlNet\footnote{\url{https://huggingface.co/diffusers/controlnet-canny-sdxl-1.0}} and depth-based ControlNet\footnote{\url{https://huggingface.co/diffusers/controlnet-depth-sdxl-1.0}}. The $\alpha$ in the rigid object, cloth, and liquid experiments are set to 0.9, 0.75, and 0.4, respectively. We use the DDIM sampler \cite{song2021denoising} with classifier-free guidance \cite{ho2022classifier} and 50 sampling steps in our experiments on one NVIDIA A6000 GPU. The version of the Blender is 3.6. We generate 80-frame sequences of edge and depth maps at a resolution of $1920 \times 1080$ for each prompt. Theoretically, our method can generate motion video of any length and resolution. For conciseness, in this paper, we show the cropped video with $1080 \times 1080$ resolution. By the way, the videos in this experimental section may look slow, which is because too many videos are displayed at the same time on the same page. To view the motion in these videos, please use Acrobat Reader\footnote{\url{https://www.adobe.com/acrobat/pdf-reader.html}}. The original videos can be found in our supplementary material.

\def\basketballsize{0.15}
\begin{figure}[tbp]
	\centering
	\setlength{\abovecaptionskip}{-0.15cm}
	\begin{subfigure}[b]{0.16\textwidth}
		\centering
		\includegraphics[width=0.99\linewidth]{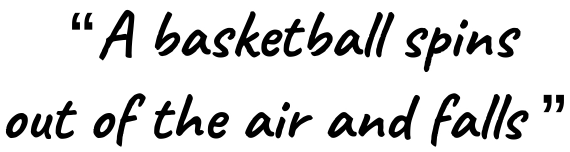}\\
		\quad
	\end{subfigure}%
	\begin{subfigure}[b]{0.16\textwidth}
		\centering
		\includegraphics[width=0.88\linewidth]{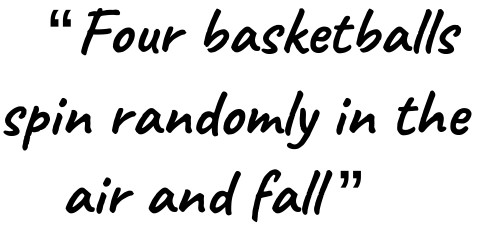}
	\end{subfigure}%
	\begin{subfigure}[b]{0.16\textwidth}
		\centering
		\includegraphics[width=0.88\linewidth]{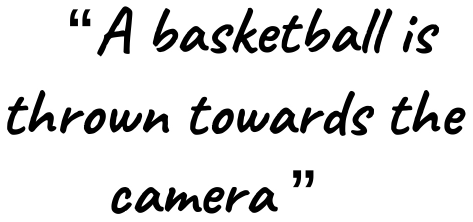}
	\end{subfigure}%
	\hfill
	\begin{subfigure}[b]{0.16\textwidth}
		\animategraphics[width=0.99\linewidth, autoplay, loop]{24}{video_imgs/formal_experiment/basketball/video-A-basketball-spins-out-of-the-air-and-falls_crop_compress_35/}{0}{\maxzhen}
  \phantomcaption
		\label{sec4:basketball_1}
	\end{subfigure}%
	\begin{subfigure}[b]{0.16\textwidth}
		\animategraphics[width=0.99\linewidth, autoplay, loop]{24}{video_imgs/formal_experiment/basketball/video-Four/}{0}{\maxzhen}
  \phantomcaption
		\label{sec4:basketball_2}
	\end{subfigure}%
	\begin{subfigure}[b]{0.16\textwidth}
		\animategraphics[width=0.99\linewidth, autoplay, loop]{7}{video_imgs/formal_experiment/basketball/basket_fall_camera_crop_compress_35/}{0}{17}
  \phantomcaption
		\label{sec4:basketball_3}
	\end{subfigure}%
	\caption{\methodshort's results on basketball drop and collision. \emph{Best viewed with \href{https://www.adobe.com/acrobat/pdf-reader.html}{Acrobat Reader} for animation. } }
	\label{sec4:basketball}
	\vspace{0.35cm}
	\centering
		\setlength{\abovecaptionskip}{-0.15cm}
	\begin{subfigure}[b]{0.16\textwidth}
		\centering
		\includegraphics[width=0.95\linewidth]{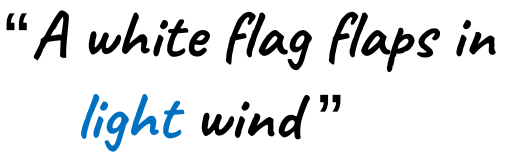}
	\end{subfigure}%
	\begin{subfigure}[b]{0.16\textwidth}
		\centering
		\includegraphics[width=0.95\linewidth]{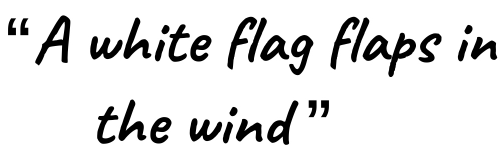}
	\end{subfigure}%
	\begin{subfigure}[b]{0.16\textwidth}
		\centering
		\includegraphics[width=0.95\linewidth]{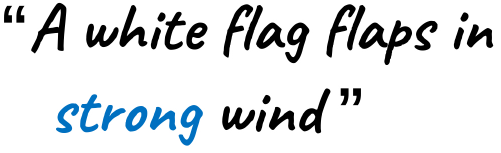}
	\end{subfigure}%
	\hfill
	\begin{subfigure}[b]{0.16\textwidth}
		\animategraphics[width=0.99\linewidth, autoplay, loop]{24}{video_imgs/formal_experiment/Flag/video-flag-light/}{0}{\maxzhen}
		\label{sec4:flag_1}
	\end{subfigure}%
	\begin{subfigure}[b]{0.16\textwidth}
		\animategraphics[width=0.99\linewidth, autoplay, loop]{24}{video_imgs/formal_experiment/Flag/video-flag/}{0}{\maxzhen}
		\label{sec4:flag_2}
	\end{subfigure}%
	\begin{subfigure}[b]{0.16\textwidth}
		\animategraphics[width=0.99\linewidth, autoplay, loop]{24}{video_imgs/formal_experiment/Flag/video-flag-fierce-wind_crop_compress_35/}{0}{\maxzhen}
		\label{sec4:flag_3}
	\end{subfigure}%
	\caption{\methodshort's results on a fluttering flag.}
	 \label{sec4:flag}
	 \vspace{0.35cm}
    \centering
    	\setlength{\abovecaptionskip}{-0.15cm}
    \begin{subfigure}[b]{0.16\textwidth}
    	\centering
    	\includegraphics[width=0.94\linewidth]{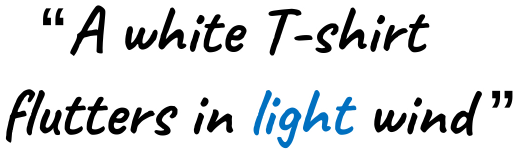}
    \end{subfigure}%
    \begin{subfigure}[b]{0.16\textwidth}
    	\centering
    	\includegraphics[width=0.90\linewidth]{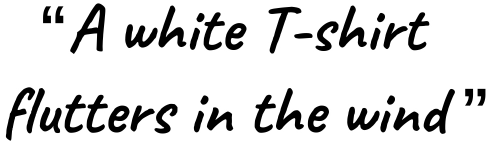}
    \end{subfigure}%
    \begin{subfigure}[b]{0.16\textwidth}
    	\centering
    	\includegraphics[width=\linewidth]{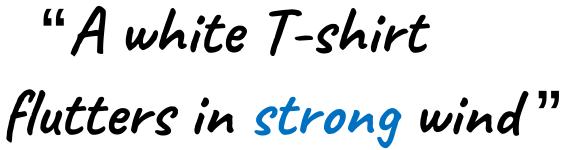}
    \end{subfigure}%
    \hfill
    \begin{subfigure}[b]{0.16\textwidth}
        \animategraphics[width=0.99\linewidth, autoplay, loop]{24}{video_imgs/formal_experiment/T-shirt/video-light/}{0}{\maxzhen}
        \label{sec4:T-shirt_1}
    \end{subfigure}%
    \begin{subfigure}[b]{0.16\textwidth}
        \animategraphics[width=0.99\linewidth, autoplay, loop]{24}{video_imgs/formal_experiment/T-shirt/video-wind/}{0}{\maxzhen}
        \label{sec4:T-shirt_2}
    \end{subfigure}%
        \begin{subfigure}[b]{0.16\textwidth}
       \animategraphics[width=0.99\linewidth, autoplay, loop]{24}{video_imgs/formal_experiment/T-shirt/video-fierce/}{0}{80}
        \label{sec4:T-shirt_3}
    \end{subfigure}%
     \caption{\methodshort's results on a fluttering T-shirt.}
     \label{sec4:T-shirt}

\end{figure}

\subsection{Controlling Physical Properties}
We demonstrate the generative capabilities of our method across three physical scenarios, highlighting how it enables control over specific physical properties through user prompts to influence the overall generation results.
\paragraph{Basketball Drop and Collision.} Figure \ref{sec4:basketball} displays basketball motion videos generated by our method with three prompts. In Figure \ref{sec4:basketball} (left), the basketball maintains a high degree of realism in its texture while spinning, and accurately replicates the bouncing behavior after collision with the floor. Figure \ref{sec4:basketball} (middle) demonstrates that our method can precisely control the number of basketballs and efficiently generate the collisions and bounces that occur when multiple basketballs land.
Impressively, as shown in Figure \ref{sec4:basketball} (right), when the user requests that the basketball is thrown towards the camera, GPT-4 calculates the necessary initial velocity of the basketball based on its fall time in the generated script, thereby achieving a visually convincing effect. This demonstrates that our approach can be combined with the physical knowledge that GPT-4 has to control the content of the video generation (see the supplementary material for more details).
\vspace{-0.3cm}
\paragraph{Cloth Fluttering in Wind.} Figures \ref{sec4:flag} and \ref{sec4:T-shirt} validate our method's capability in generating the motion of cloth objects influenced by wind. Utilizing existing physics engines for simulation, \methodshort{} generates the fluctuations and waves of cloth under different wind strengths. In Figure \ref{sec4:flag}, we present the generated results of a flag fluttering. The flag exhibits complex ripple and wave patterns under different wind strengths. Figure \ref{sec4:T-shirt} shows the motion of an irregular cloth object, T-shirt, under different wind strengths. Influenced by the physical properties of the fabric, such as elasticity and weight, the T-shirt undergoes flapping and twisting, with visible changes in creases and wrinkles.
\vspace{-0.3cm}

\paragraph{Water Pouring into a Mug.} Figure \ref{sec4:water} shows three videos of water of different viscosities being poured into a mug. When the viscosity is low, the flowing water collides and merges with the water in the mug, creating complex turbulence on the surface. As the viscosity increases, the flow becomes slower and the water begins to stick together.

\subsection{Comparisons with Baselines}
\begin{figure}[tbp]
    \centering
    \setlength{\abovecaptionskip}{-0.15cm}
    \begin{subfigure}[b]{0.16\textwidth}
    	\centering
    	\includegraphics[width=0.80\linewidth]{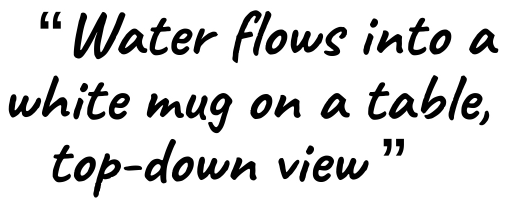}
    \end{subfigure}%
    \begin{subfigure}[b]{0.16\textwidth}
    	\centering
    	\includegraphics[width=0.85\linewidth]{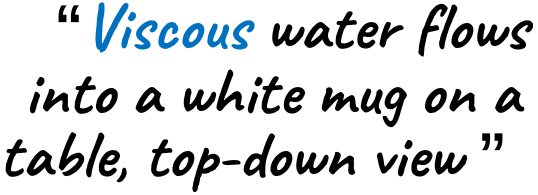}
    \end{subfigure}%
    \begin{subfigure}[b]{0.16\textwidth}
    	\centering
    	\includegraphics[width=0.95\linewidth]{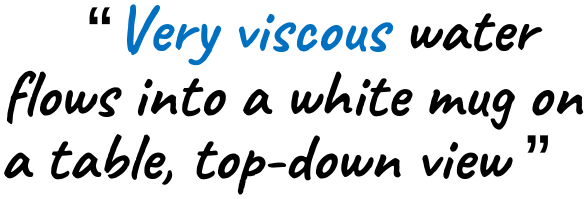}
            
    \end{subfigure}%
    \hfill
    \begin{subfigure}[b]{0.16\textwidth}
        \animategraphics[width=0.99\linewidth, autoplay, loop]{24}{video_imgs/formal_experiment/water/normal-water/}{0}{\maxzhen}
        \label{sec4:water_1}
    \end{subfigure}%
    \begin{subfigure}[b]{0.16\textwidth}
        \animategraphics[width=0.99\linewidth, autoplay, loop]{24}{video_imgs/formal_experiment/water/viscous-water/}{0}{\maxzhen}
        \label{sec4:water_2}
    \end{subfigure}%
        \begin{subfigure}[b]{0.16\textwidth}
       \animategraphics[width=0.99\linewidth, autoplay, loop]{24}{video_imgs/formal_experiment/water/very-viscous-water/}{0}{\maxzhen}
        \label{sec4:water_3}
    \end{subfigure}%
     \caption{\methodshort's results on the water pouring. \emph{Best viewed with \href{https://www.adobe.com/acrobat/pdf-reader.html}{Acrobat Reader} for animation. }}
     \label{sec4:water}
\vspace{-2mm}
\end{figure}

We compare our GPT4Motion against four baselines:
1) \textit{AnimateDiff} \cite{guo2023animatediff}, which combines Stable Diffusion with a motion module, augmented by Realistic Vision DreamBooth\footnote{\url{https://civitai.com/models/4201?modelVersionId=29460}};
2) \textit{ModelScope} \cite{wang2023modelscope}, incorporating spatial-temporal convolution and attention mechanisms into Stable Diffusion for T2V tasks;
3) \textit{Text2Video-Zero} \cite{khachatryan2023text2video}, which leverages Stable Diffusion's image-to-image capabilities for generating videos through cross-attention and modified latent code sampling;
4) \textit{DirecT2V} \cite{hong2023large}, employing a LLM for frame-level descriptions from prompts, with rotational value mapping and dual-softmax for continuity.
To maintain the size of the paper, we only compare GPT4Motion with these baselines on three examples. More comparisons are given in the supplementary material.
\vspace{-0.3cm}
\paragraph{A Basketball Free Falls in the Air.} As shown in Figure \ref{sec1:comparison_experiment}, the baselines' results do not match the user prompt. DirecT2V and Text2Video-Zero face challenges in texture realism and motion consistency, whereas AnimateDiff and ModelScope improve video smoothness but struggle with consistent textures and realistic movements. In contrast to these methods, \methodshort{} can generate smooth texture changes during the falling of the basketball, and bouncing after collision with the floor, which appear more realistic.
\vspace{-0.3cm}
\paragraph{A White Flag Flaps in the Wind.} As shown in Figure \ref{sec4:compared_baseline} (1st row), the videos generated by AnimateDiff and Text2Video-Zero exhibit artifacts/distortions in the flags, whereas ModelScope and DirecT2V are unable to smoothly generate the gradual transition of flag fluttering in the wind. However, as shown in the middle of Figure \ref{sec4:flag}, the video generated by \methodshort{} can show the continuous change of wrinkles and ripples on the flag under the effect of gravity and wind.
\vspace{-0.3cm}
\paragraph{Water Flows into a White Mug on a Table, Top-Down View.} As shown in Figure \ref{sec4:compared_baseline} (2nd row), all the baselines' results fail to align with the user prompt. While the videos from AnimateDiff and ModelScope reflect changes in the water flow, they cannot capture the physical effects of water pouring into a mug. The videos generated by Text2Video-Zero and DirecT2V, on the other hand, show a constantly jittering mug. In comparison, as shown in Figure \ref{sec4:water} (left), \methodshort{} generates the video that accurately depicts the surge of water as it collides with the mug, offering a more realistic effect.

\begin{table}
\scriptsize
	 \centering
	 \setlength{\tabcolsep}{0.1mm}{
	\begin{tabular}{c c c c}
		\multicolumn{4}{c}{\includegraphics[width=0.6\linewidth]{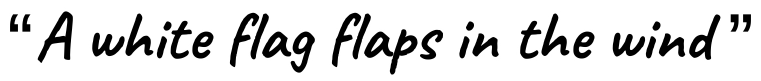}}\\
		\animategraphics[width=0.25\linewidth, autoplay, loop]{7}{video_imgs/comparison_experiment/flag-AnimateDiff/}{0}{15}& 
		\animategraphics[width=0.25\linewidth, autoplay, loop]{7}{video_imgs/comparison_experiment/flag-ModelScope/}{0}{15} &
		\animategraphics[width=0.25\linewidth, autoplay, loop]{7}{video_imgs/comparison_experiment/flag-Text2Video/}{0}{7} &
		\animategraphics[width=0.25\linewidth, autoplay, loop]{7}{video_imgs/comparison_experiment/flag-direcT2V/}{0}{7}\\
  \multicolumn{1}{c}{AnimateDiff}&ModelScope &Text2Video-Zero &DirecT2V\\
		
		\multicolumn{4}{c}{\includegraphics[width=0.62\linewidth]{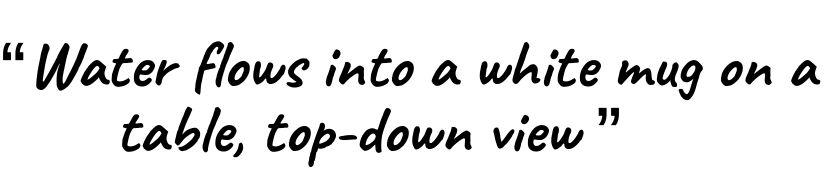}}\\
		\animategraphics[width=0.25\linewidth, autoplay, loop]{7}{video_imgs/comparison_experiment/water-AnimateDiff/}{0}{15}&
		\animategraphics[width=0.25\linewidth, autoplay, loop]{7}{video_imgs/comparison_experiment/water-ModelScope/}{0}{15}&
		\animategraphics[width=0.25\linewidth, autoplay, loop]{7}{video_imgs/comparison_experiment/water-Text2Video/}{0}{7}&
		\animategraphics[width=0.25\linewidth, autoplay, loop]{7}{video_imgs/comparison_experiment/water-direcT2V/}{0}{7}\\
		
		\multicolumn{1}{c}{AnimateDiff}&ModelScope &Text2Video-Zero &DirecT2V\\
	\end{tabular}
}
	\vspace{-0.15cm}
	\captionof{figure}{Videos generated by four text-to-video baselines with two user prompts.}
        \label{sec4:compared_baseline}
\end{table}

\begin{table}[t]
    \centering
    \resizebox{0.45\textwidth}{!}{
    \begin{tabular}{cccccc}
    \toprule
    Method & Motion$\uparrow$ & CLIP$\uparrow$ & Flickering$\uparrow$ \\ 
    \midrule
    GPT4Motion & \textbf{0.993 $\pm$ 0.003} & \textbf{0.260 $\pm$ 0.022} & \textbf{0.990 $\pm$ 0.006} \\ 
    AnimateDiff & 0.991 $\pm$ 0.002   & 0.257 $\pm$ 0.020 & 0.988 $\pm$ 0.002 \\ 
    ModelScope & 0.937 $\pm$ 0.051 & 0.252 $\pm$ 0.036 & 0.924 $\pm$ 0.059 \\ 
    Text2Video-Zero & 0.946 $\pm$ 0.015 & 0.252 $\pm$ 0.024 & 0.928 $\pm$ 0.009 \\ 
    DirecT2V & 0.879 $\pm$ 0.067 & 0.253 $\pm$ 0.033 & 0.870 $\pm$ 0.071 \\ 
    \bottomrule
    \end{tabular}
    }
    \caption{Quantitative comparison across various methods. The best performances are denoted in bold.}
    \label{tab:video_generation_comparison}
    \vspace{-5mm}
\end{table}

\vspace{-2mm}
\paragraph{Quantitative Evaluation and User Study.} 
We select three metrics for quantitative comparisons: Motion Smoothness \cite{huang2023vbench}, which represents the fluidity of video motion and reflects the physical accuracy to some extent; CLIP scores \cite{liu2023evalcrafter}, indicative of the alignment between the prompt and the video; and Temporal Flickering \cite{huang2023vbench}, which illustrates the flickering level of the generated videos. Please refer to the supplementary material for details on each metric. The results, as shown in Table \ref{tab:video_generation_comparison}, demonstrate that our GPT4Motion, leveraging GPT-4 for understanding and invoking Blender to simulate physical scenes, outperforms the other four methods on all the metrics. While videos generated by GPT4Motion still exhibit some flickering, they show a significant improvement in flickering level compared to the other models. 
However, these metrics might not encompass the entire scope of video generation quality, leading us to undertake a user study for a more comprehensive evaluation. 
We also conduct a user study with 30 participants, where we show videos generated by different methods under the same prompt and ask the participants to vote for the best video based on three evaluation criteria: physical accuracy, text-video alignment, and the least amount of video flickering. Remarkably, our GPT4Motion's results obtain 100\% of the participants' votes.

\def\vspacehill{2mm}
\def\rotateboxsize{0.037}
\newcommand{\subfigspacer}{\hskip 1pt}
\def\sizefour{0.11}
\def\imagesize{0.98}
\def\minipagesize{0.11}
\begin{figure}[t]
\centering
\begin{minipage}[t]{\rotateboxsize\textwidth}
\centering
\rotatebox{90}{~~~w/o edge}
\end{minipage}%
\begin{minipage}[t]{\minipagesize\textwidth}
  \centering
  \includegraphics[width=\imagesize\linewidth]{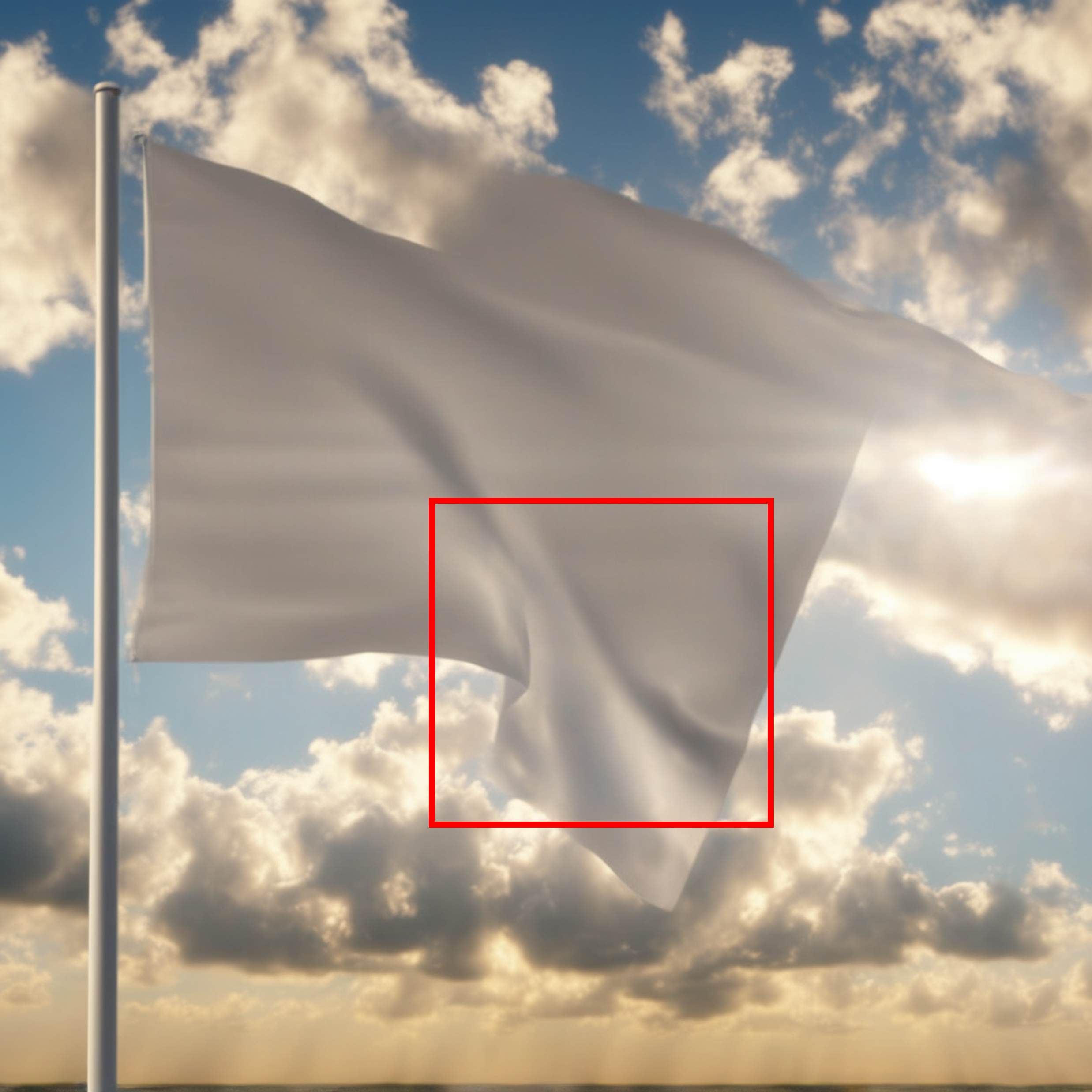}
\end{minipage}%
\begin{minipage}[t]{\minipagesize\textwidth}
  \centering
  \includegraphics[width=\imagesize\linewidth]{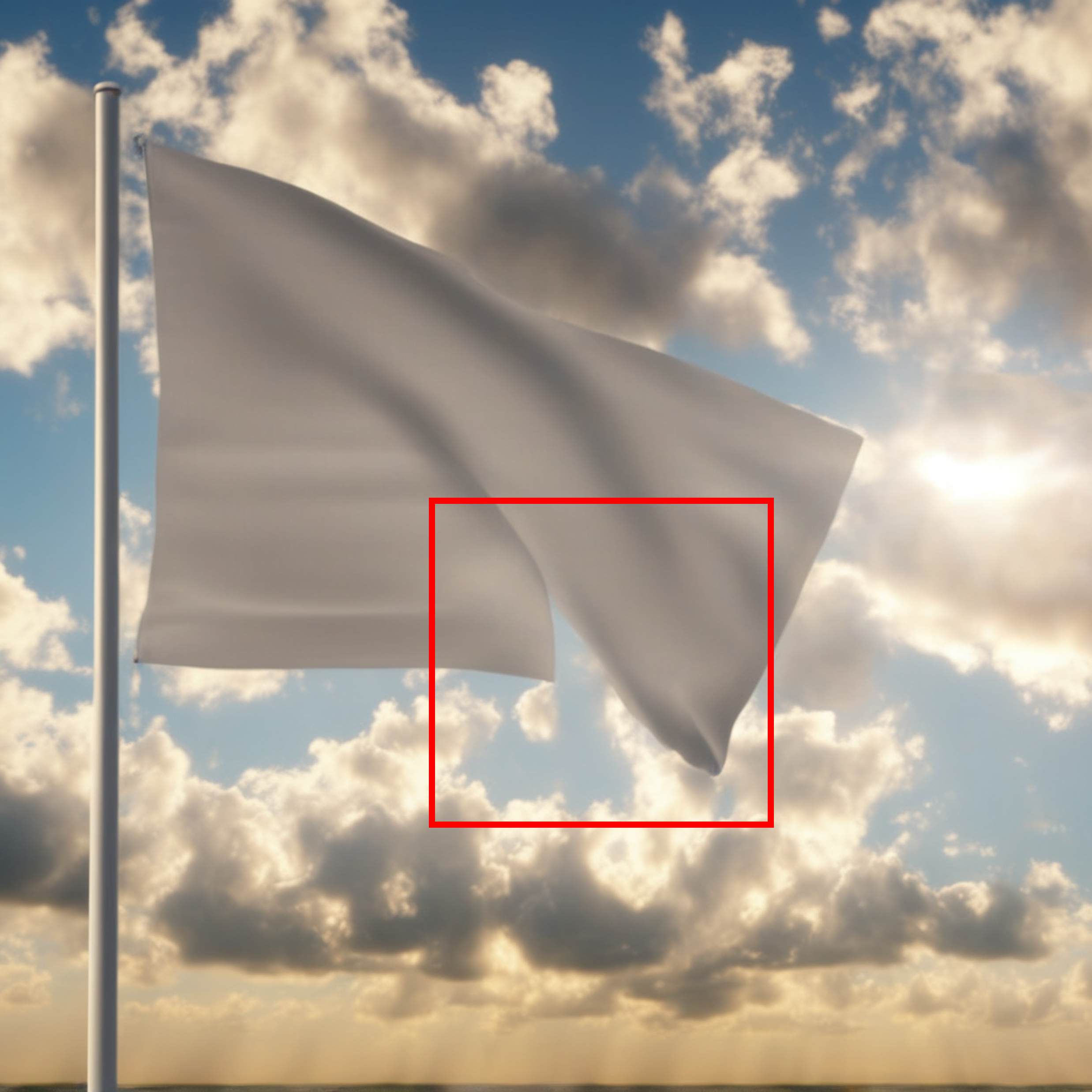}
\end{minipage}%
\begin{minipage}[t]{\minipagesize\textwidth}
  \centering
  \includegraphics[width=\imagesize\linewidth]{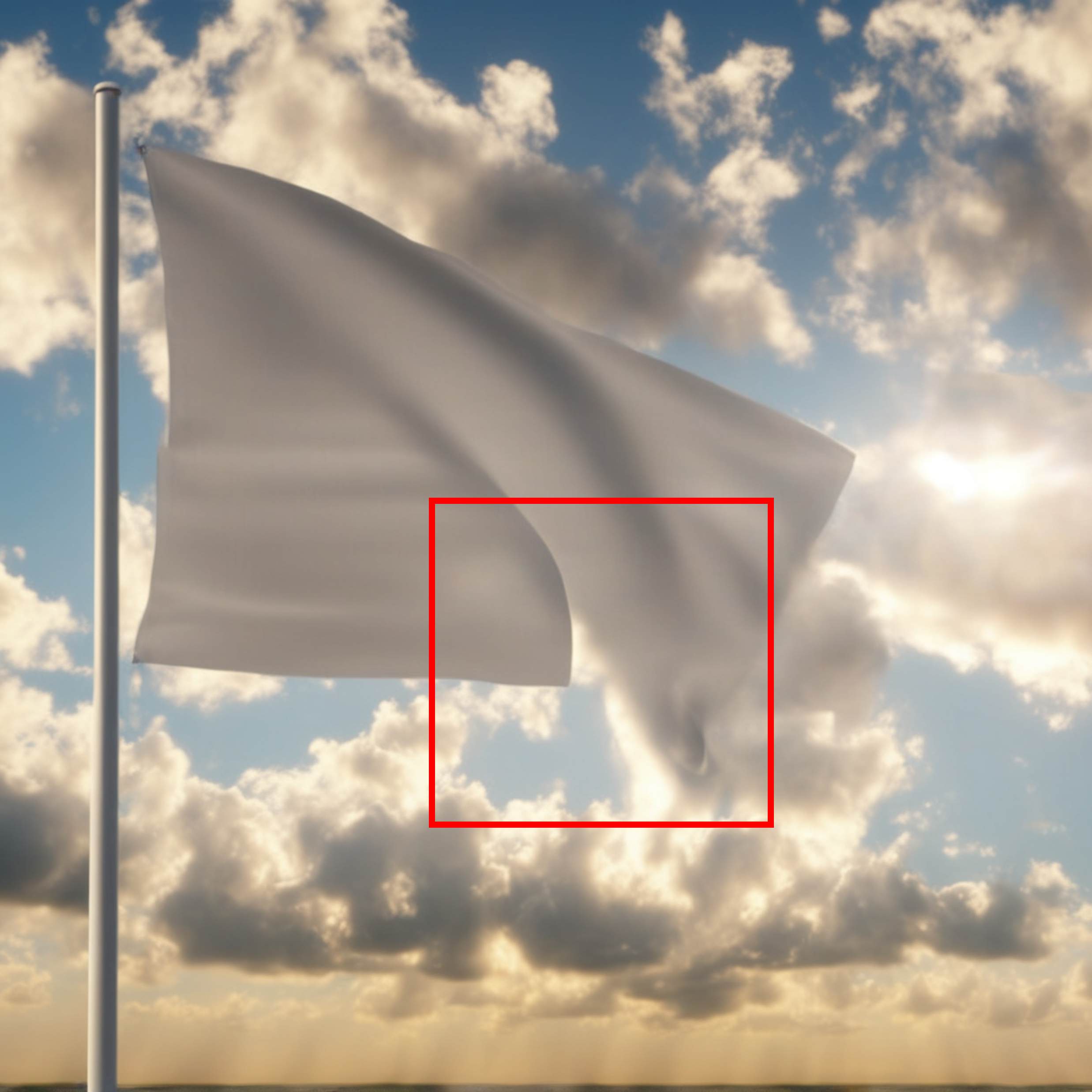}
\end{minipage}%
\begin{minipage}[t]{\minipagesize\textwidth}
  \centering
  \includegraphics[width=\imagesize\linewidth]{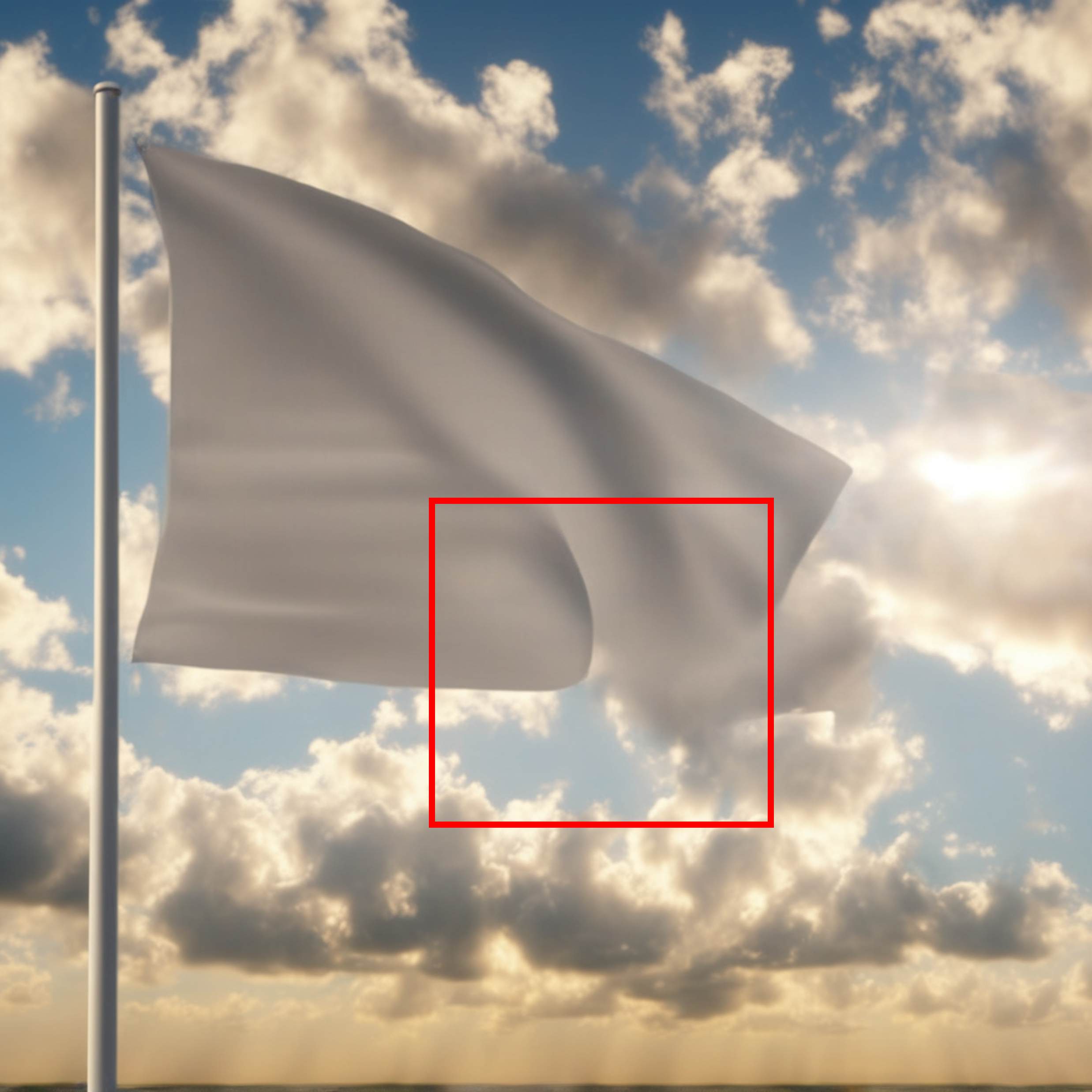}
\end{minipage}%
\hfill
\vspace{\vspacehill}
\hfill
\begin{minipage}[t]{\rotateboxsize\textwidth}
\centering
\rotatebox{90}{~~~w/o depth}
\end{minipage}%
\begin{minipage}[t]{\minipagesize\textwidth}
  \centering
  \includegraphics[width=\imagesize\linewidth]{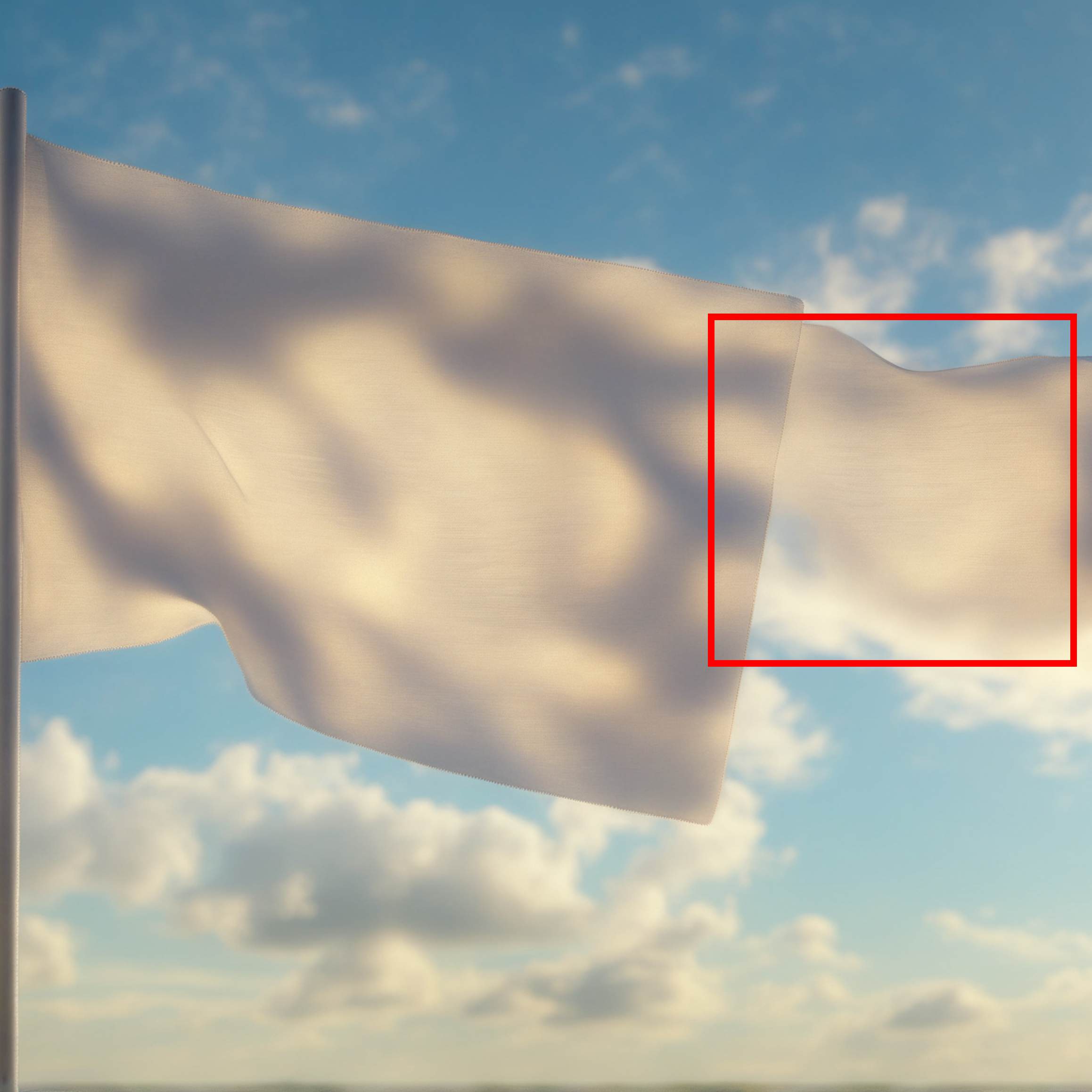}
\end{minipage}%
\begin{minipage}[t]{\minipagesize\textwidth}
  \centering
  \includegraphics[width=\imagesize\linewidth]{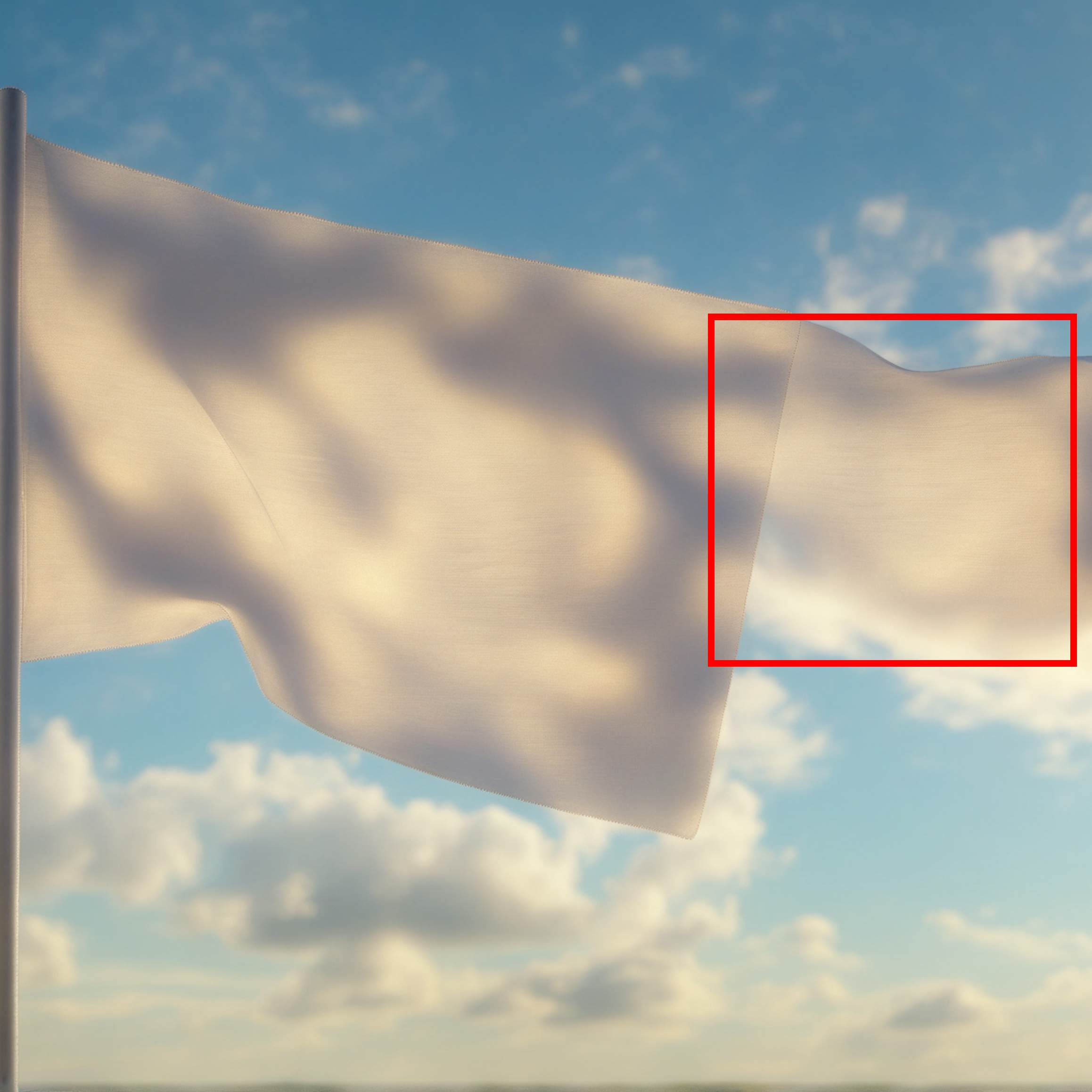}
\end{minipage}%
\begin{minipage}[t]{\minipagesize\textwidth}
  \centering
  \includegraphics[width=\imagesize\linewidth]{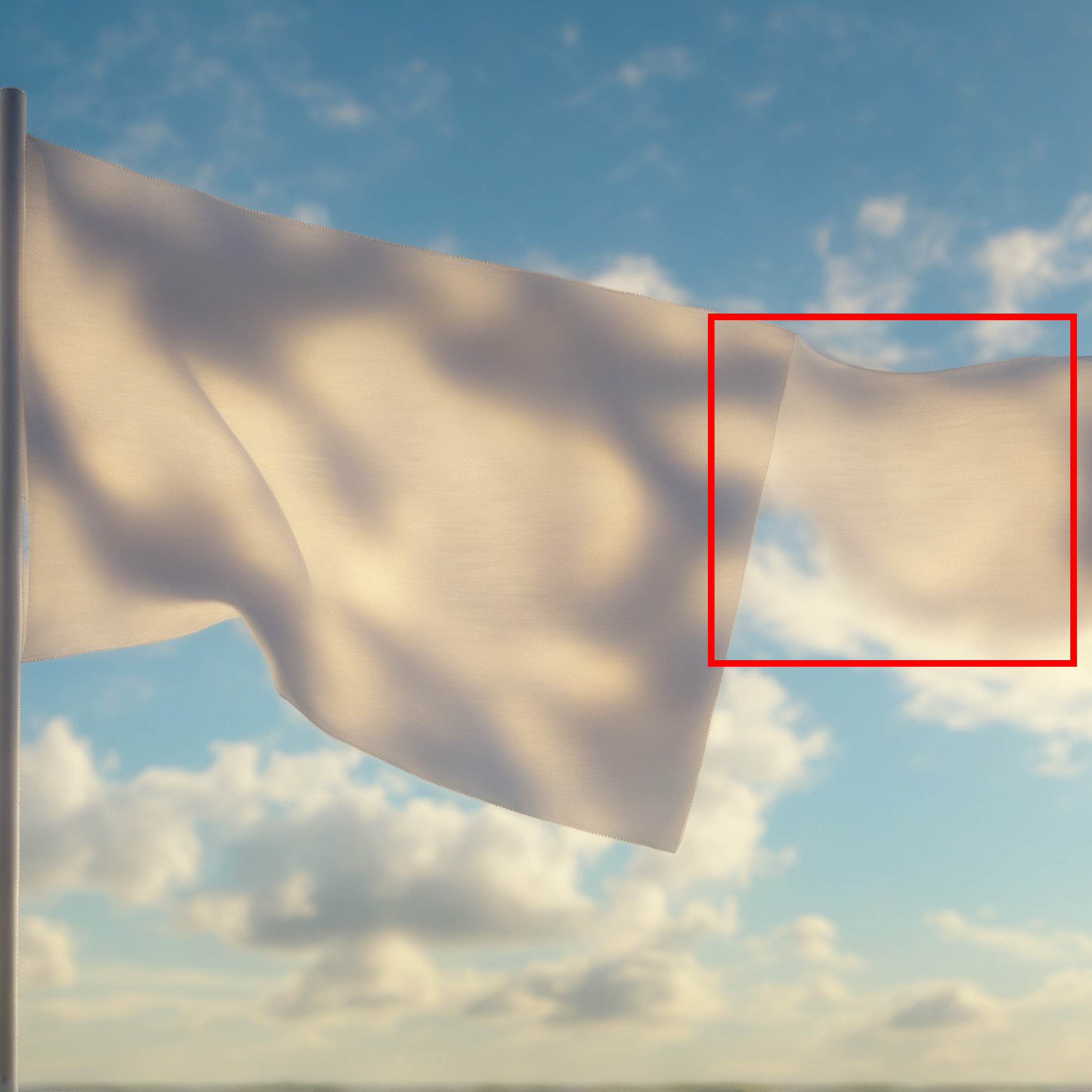}
\end{minipage}%
\begin{minipage}[t]{\minipagesize\textwidth}
  \centering
  \includegraphics[width=\imagesize\linewidth]{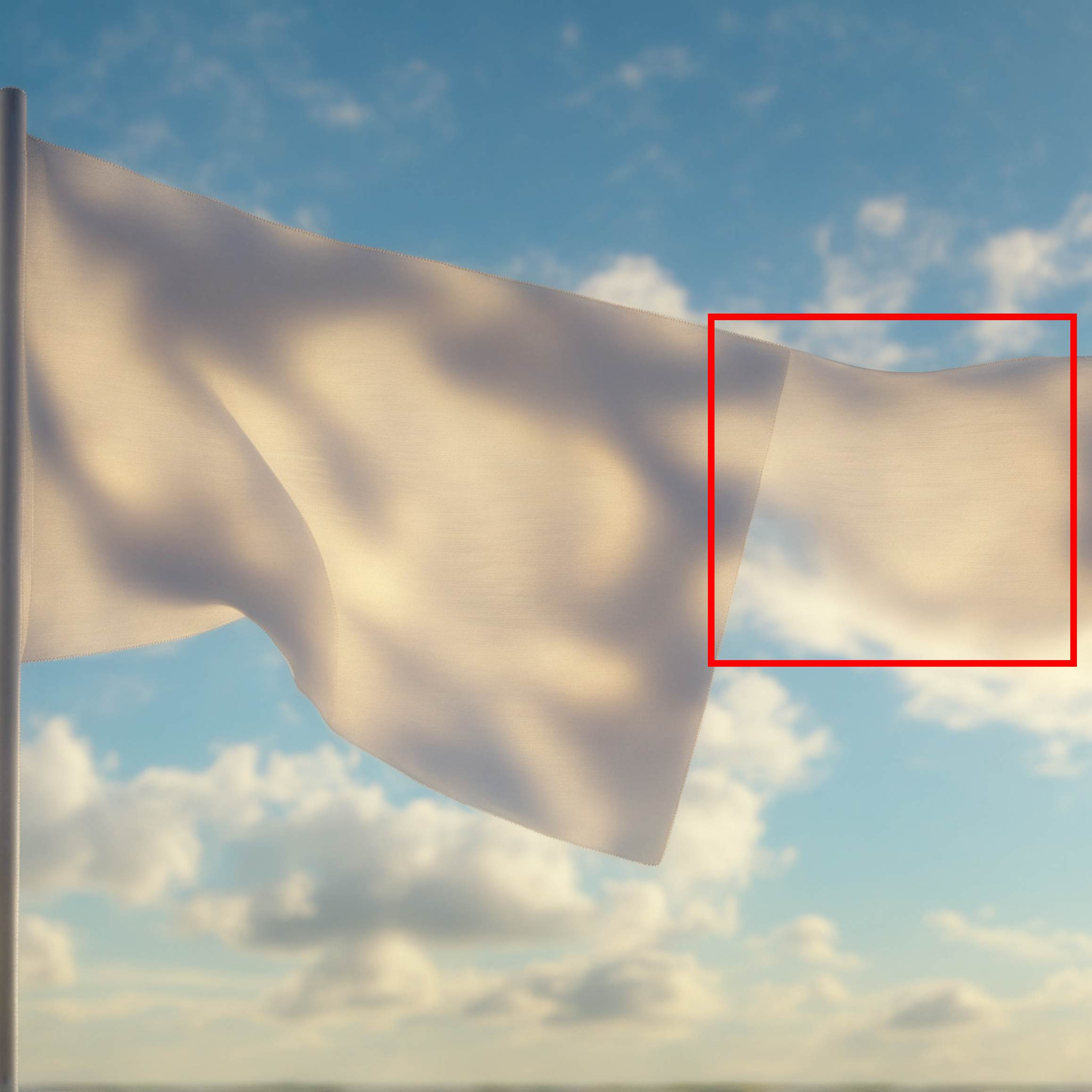}
\end{minipage}%
\hfill
\vspace{\vspacehill}
\hfill
\begin{minipage}[t]{\rotateboxsize\textwidth}
\centering
\rotatebox{90}{~~~~~~~FFA}
\end{minipage}%
\begin{minipage}[t]{\minipagesize\textwidth}
  \centering
  \includegraphics[width=\imagesize\linewidth]{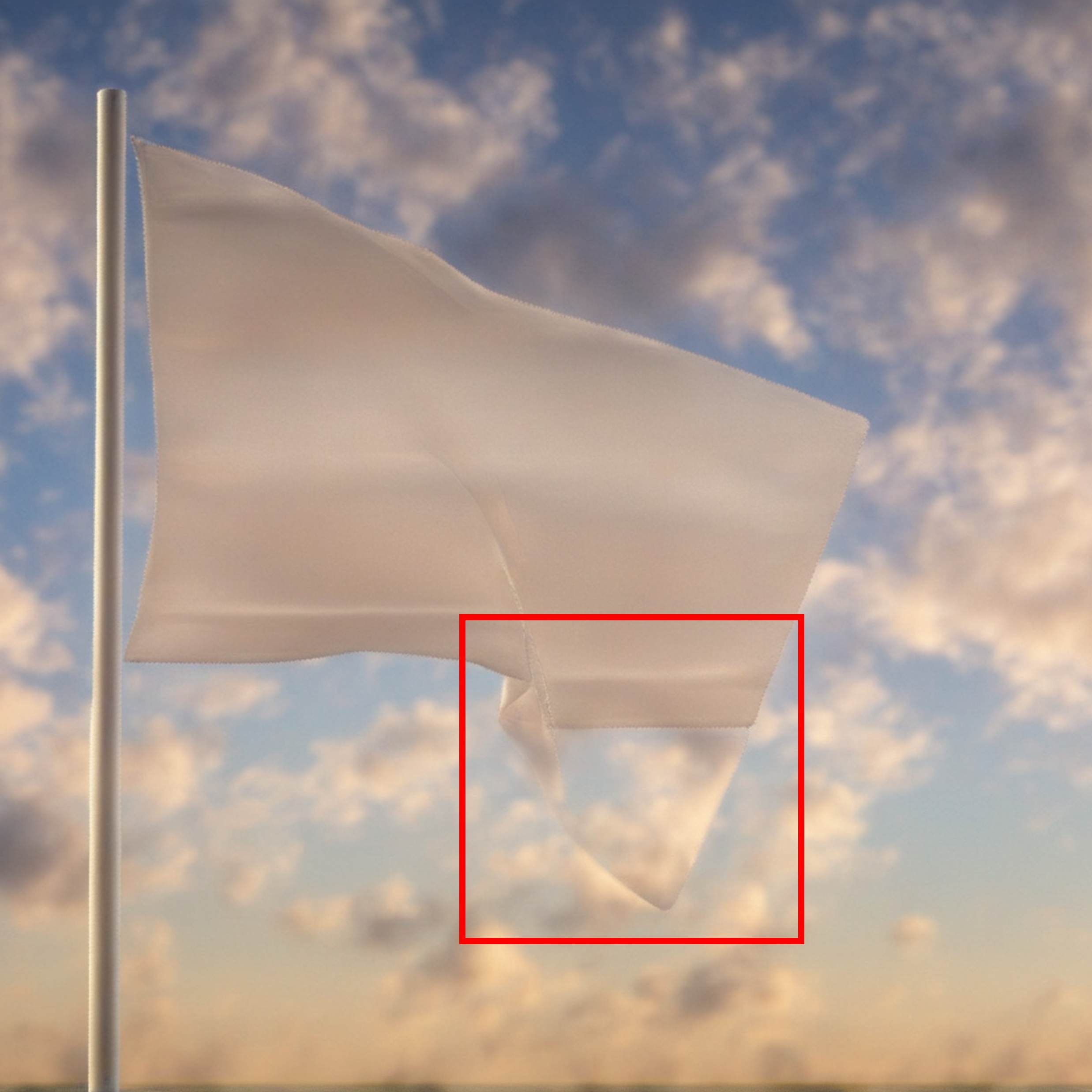}
\end{minipage}%
\begin{minipage}[t]{\minipagesize\textwidth}
  \centering
  \includegraphics[width=\imagesize\linewidth]{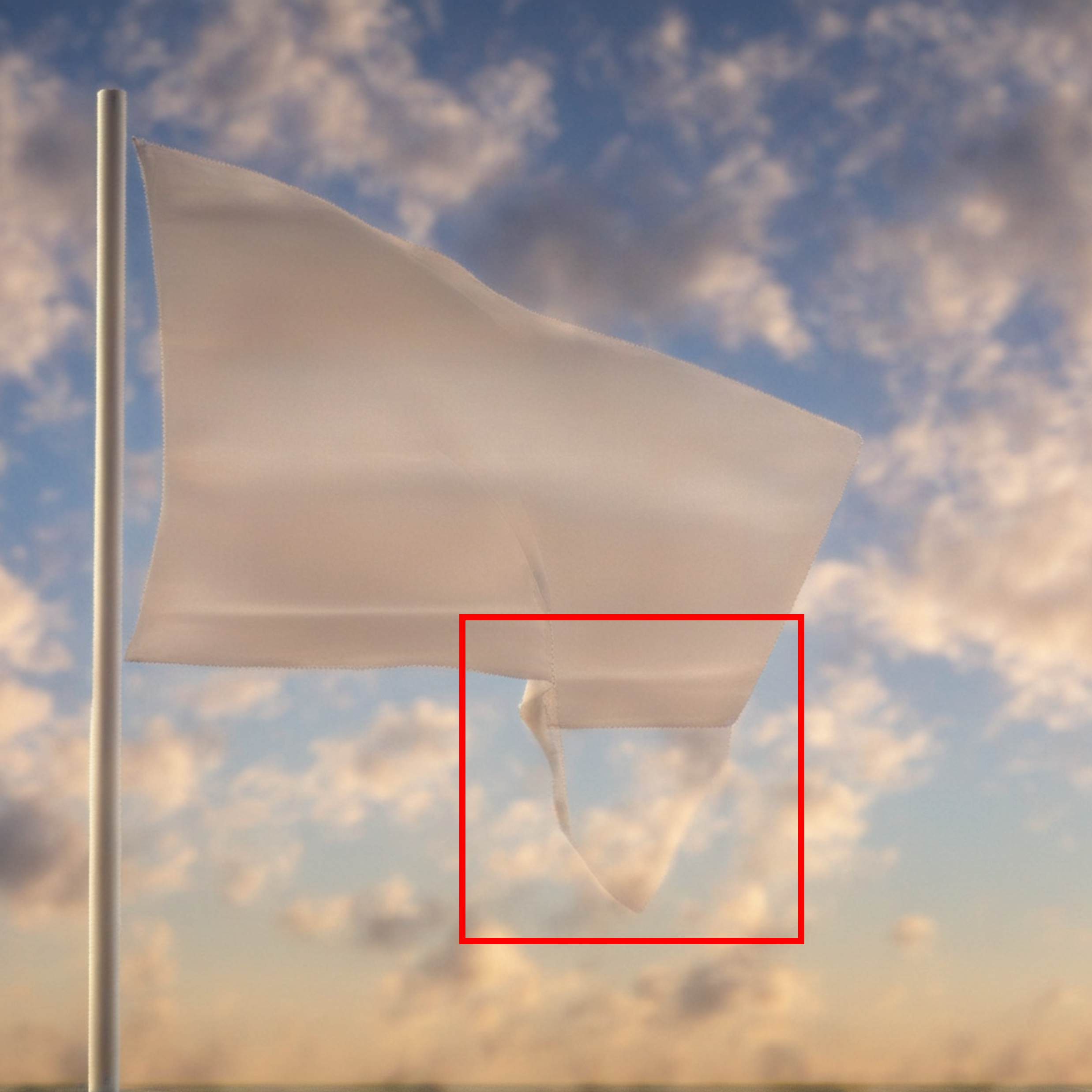}
\end{minipage}%
\begin{minipage}[t]{\minipagesize\textwidth}
  \centering
  \includegraphics[width=\imagesize\linewidth]{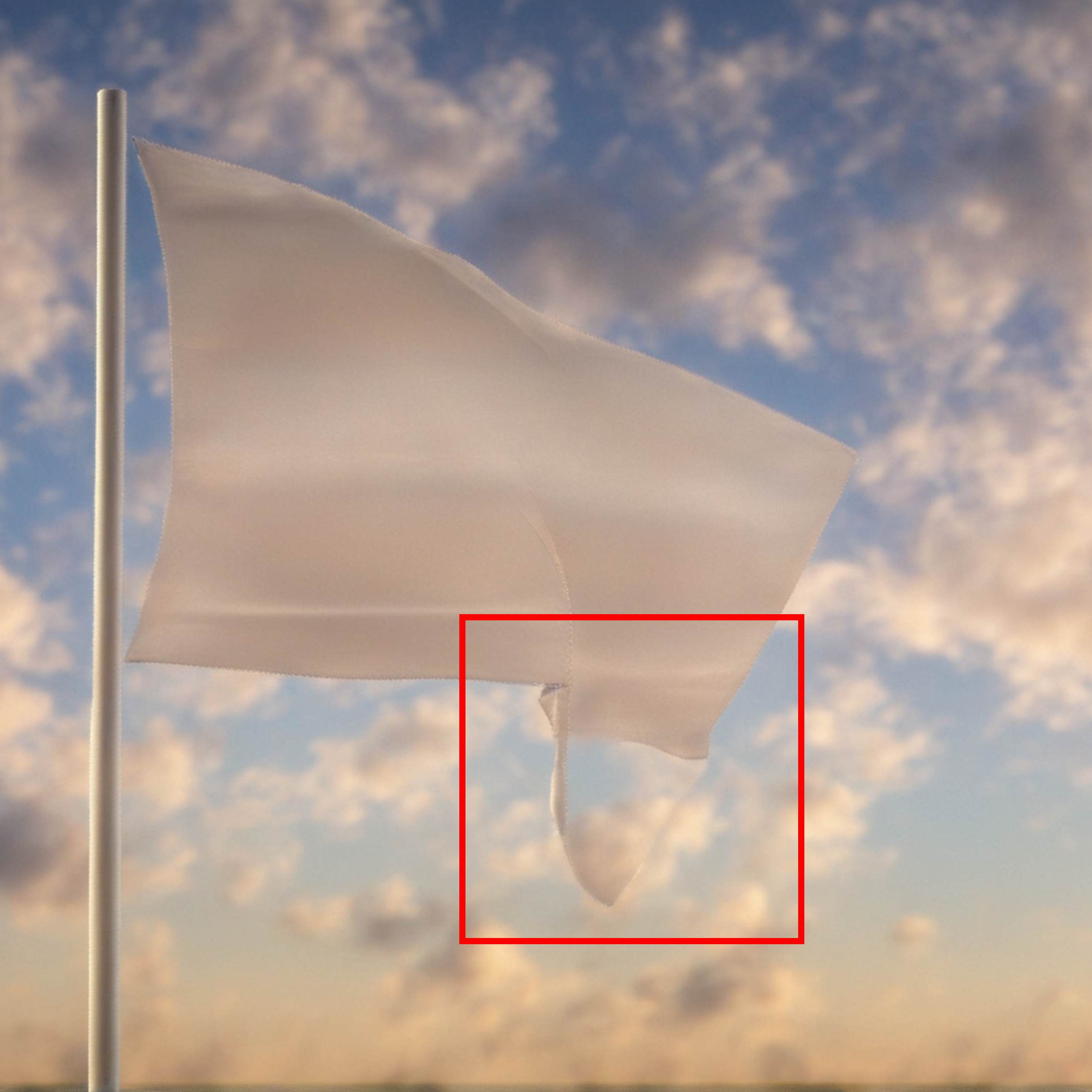}
\end{minipage}%
\begin{minipage}[t]{\minipagesize\textwidth}
  \centering
  \includegraphics[width=\imagesize\linewidth]{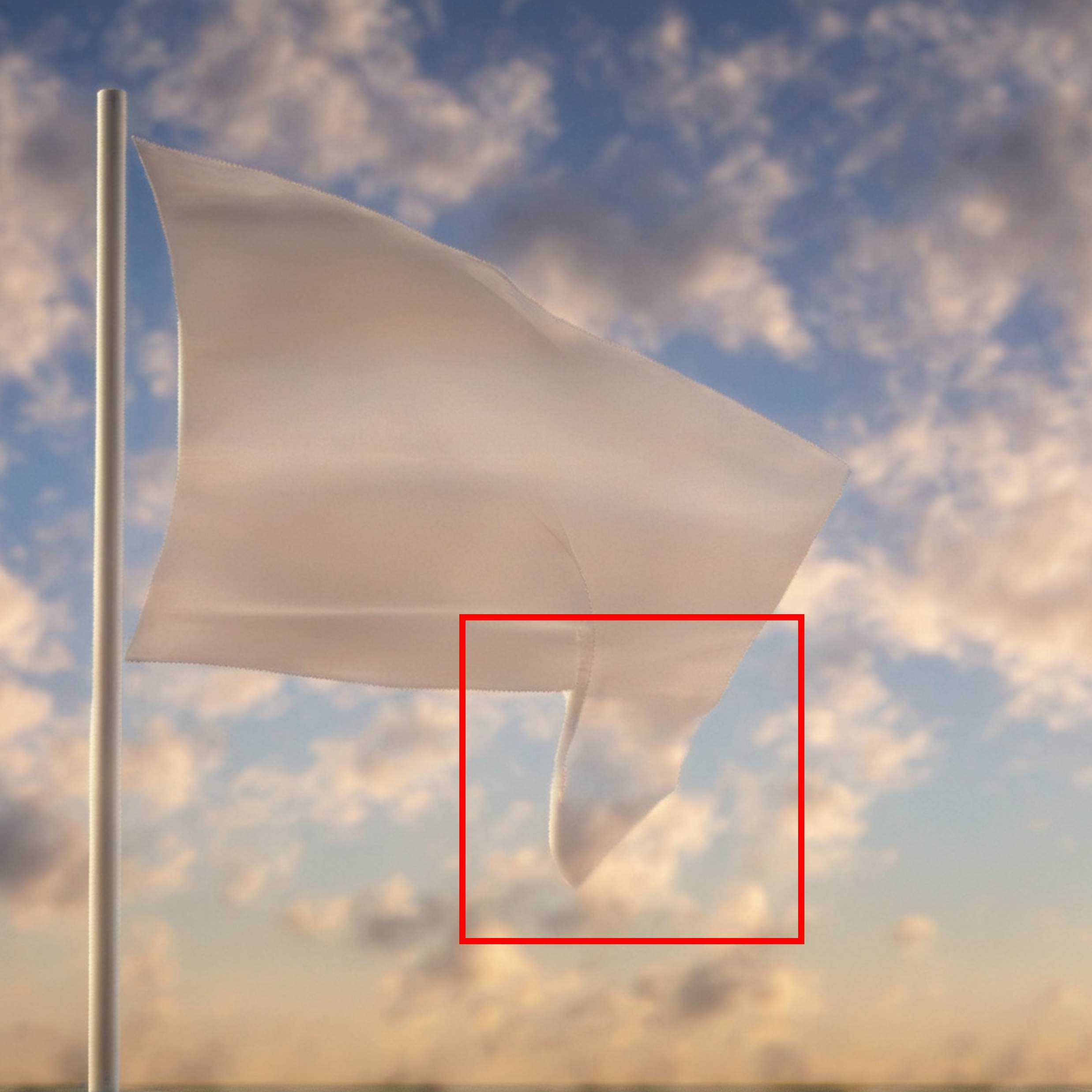}
\end{minipage}%
\hfill
\begin{minipage}[t]{\rotateboxsize\textwidth}
\centering
\rotatebox{90}{~}
\end{minipage}%
\begin{minipage}[t]{\minipagesize\textwidth}
  \centering
\small{$i$}
\end{minipage}%
\begin{minipage}[t]{\minipagesize\textwidth}
  \centering
\small{$i + 1$}
\end{minipage}%
\begin{minipage}[t]{\minipagesize\textwidth}
  \centering
\small{$i + 2$}
\end{minipage}%
\begin{minipage}[t]{\minipagesize\textwidth}
  \centering
\small{$i + 3$}
\end{minipage}%
\hfill
     \vspace{-0.1cm}
    \caption{Ablation experiments on various control conditions and cross-frame attention. Four consecutive frames are shown.}
     \label{sec4:ablation_experiment_flag_1}
\vspace{-0.1cm}
\end{figure}

\subsection{Ablation Study}
We perform an ablation study to evaluate the importance of control conditions, cross-frame attention, and $\alpha$ values in Eq.~\ref{sec3:projection}, analyzing the effect of each design separately. Experiments are conducted with the user prompt ``A white flag flaps in the wind", and the video of the complete model is shown in Figure \ref{sec4:flag} (middle). 
\paragraph{Control Conditions.} Figure \ref{sec4:ablation_experiment_flag_1} exhibits the results across frames under different controlling conditions, which shows that the model without the edge condition (\textit{w/o edge}) fails to generate correct object edges (see the first row). Additionally, the model without the depth condition (\textit{w/o depth}) not only adds extra cloth to the flag, but also mixes the flag and the cloud due to the lack of depth-of-field information.
The result of Figure \ref{sec4:flag} (middle) demonstrates that the joint use of both control conditions preserves the integrity of the object edges and well handles the problem of mixing up the flag with the sky.
\vspace{-0.4cm}
\paragraph{First-Frame Attention (FFA).} In this setting, $K_{i, 1}$ is replaced with $K_1 = W^K F_1$, and $V_{i, 1}$ is replaced with $V_1 = W^V F_1$ during the generation of the $i$-th frame in Eq.~\ref{sec3:cfa}. This means that the $i$-th frame only attends to the first frame (without paying attention to itself). As shown in Figure \ref{sec4:ablation_experiment_flag_1} (3rd row), the model \textit{FFA} results in incomplete flag generation, where part of the flag merges with the sky and white clouds. Conversely, our cross-frame attention allows the $i$-th frame during its generation to focus not only on the features of the first frame but also on its own characteristics, thereby maintaining temporal consistency and ensuring the completeness of the generated object.
\vspace{-0.4cm}
\paragraph{Different $\bm{\alpha}$ Values.}
To explore the balance of the first frame and current frame in keeping temporal consistency, we select three different $\alpha$ values for comparison. Figure \ref{sec4:ablation_experiment_flag_2} presents four generated consecutive frames. It is clear that when the $\alpha$ value is too small, the generated results suffer from distortion, while a large $\alpha$ value causes flickering (inconsistent flag color intensity). By adjusting the $\alpha$ value to an appropriate level (i.e., 0.75), the generated results maintain the fidelity of the flag and reduce the flickering.

\def\sizefour{0.12}
\def\jianxifour{0.01mm}

\def\imagesize{0.98}
\def\minipagesize{0.11}
\begin{figure}[t]
\centering
\begin{minipage}[t]{\rotateboxsize\textwidth}
\rotatebox{90}{~~~~$\alpha = 0.1$}
\end{minipage}%
\begin{minipage}[t]{\minipagesize\textwidth}
  \centering
  \includegraphics[width=\imagesize\linewidth]{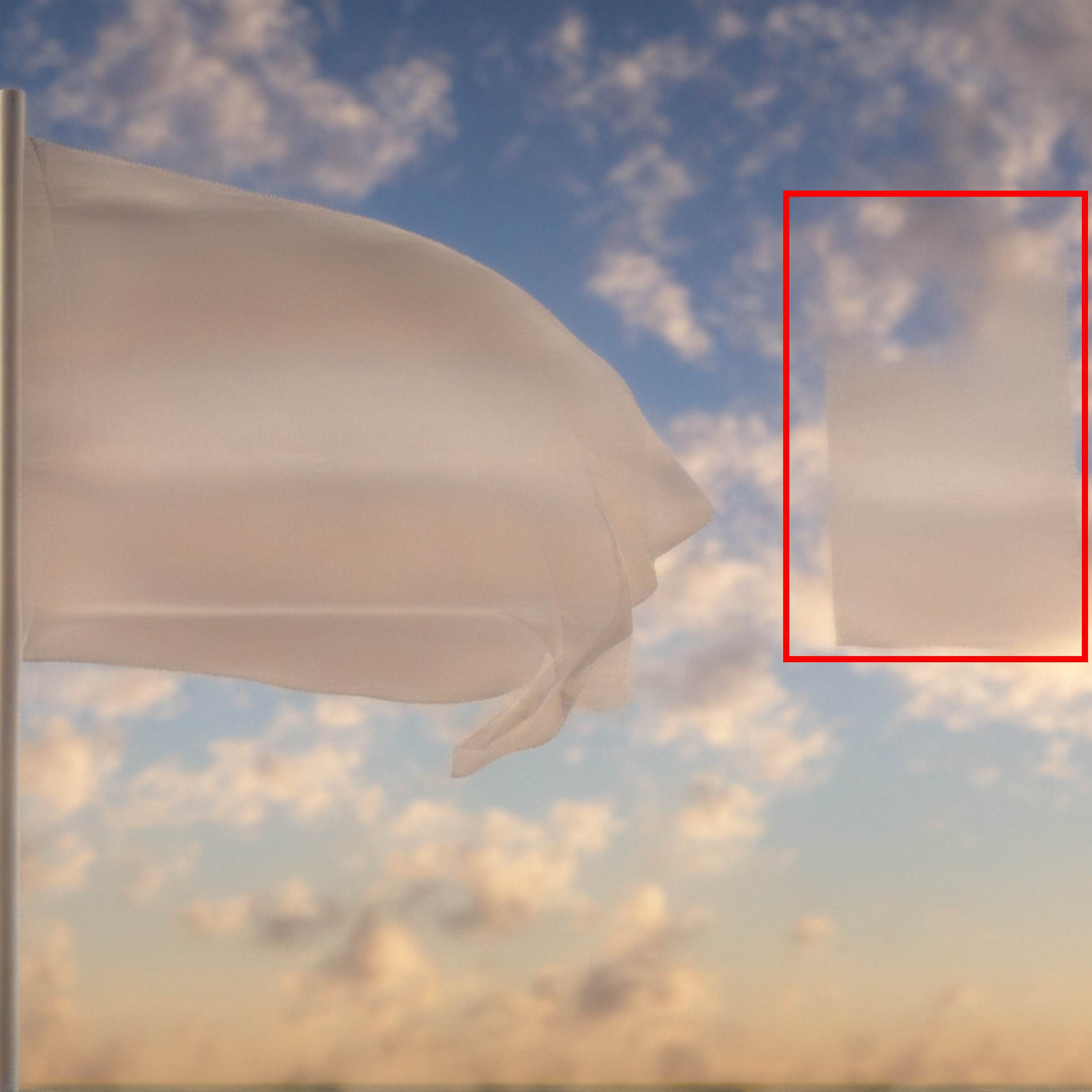}
\end{minipage}%
\begin{minipage}[t]{\minipagesize\textwidth}
  \centering
  \includegraphics[width=\imagesize\linewidth]{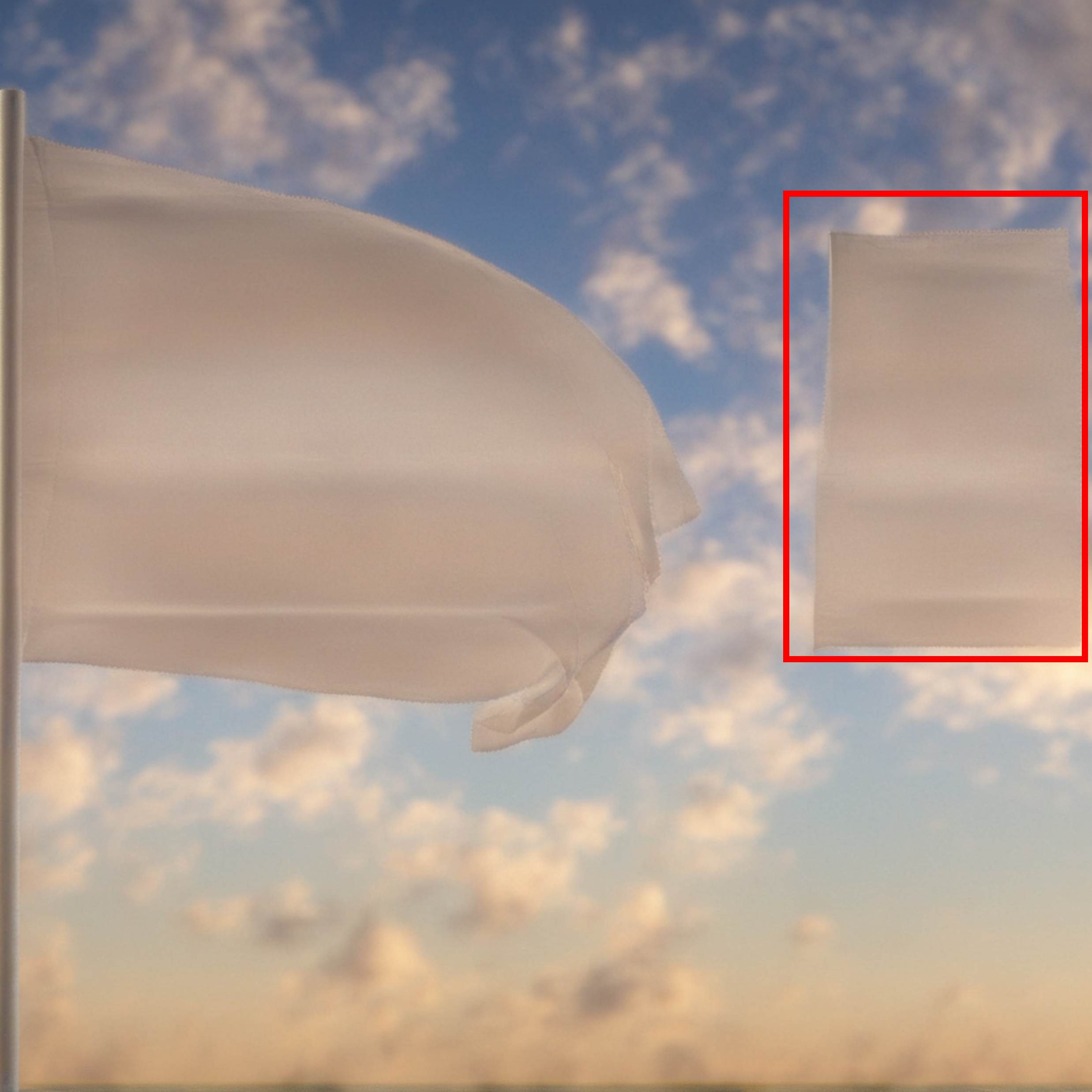}
\end{minipage}%
\begin{minipage}[t]{\minipagesize\textwidth}
  \centering
  \includegraphics[width=\imagesize\linewidth]{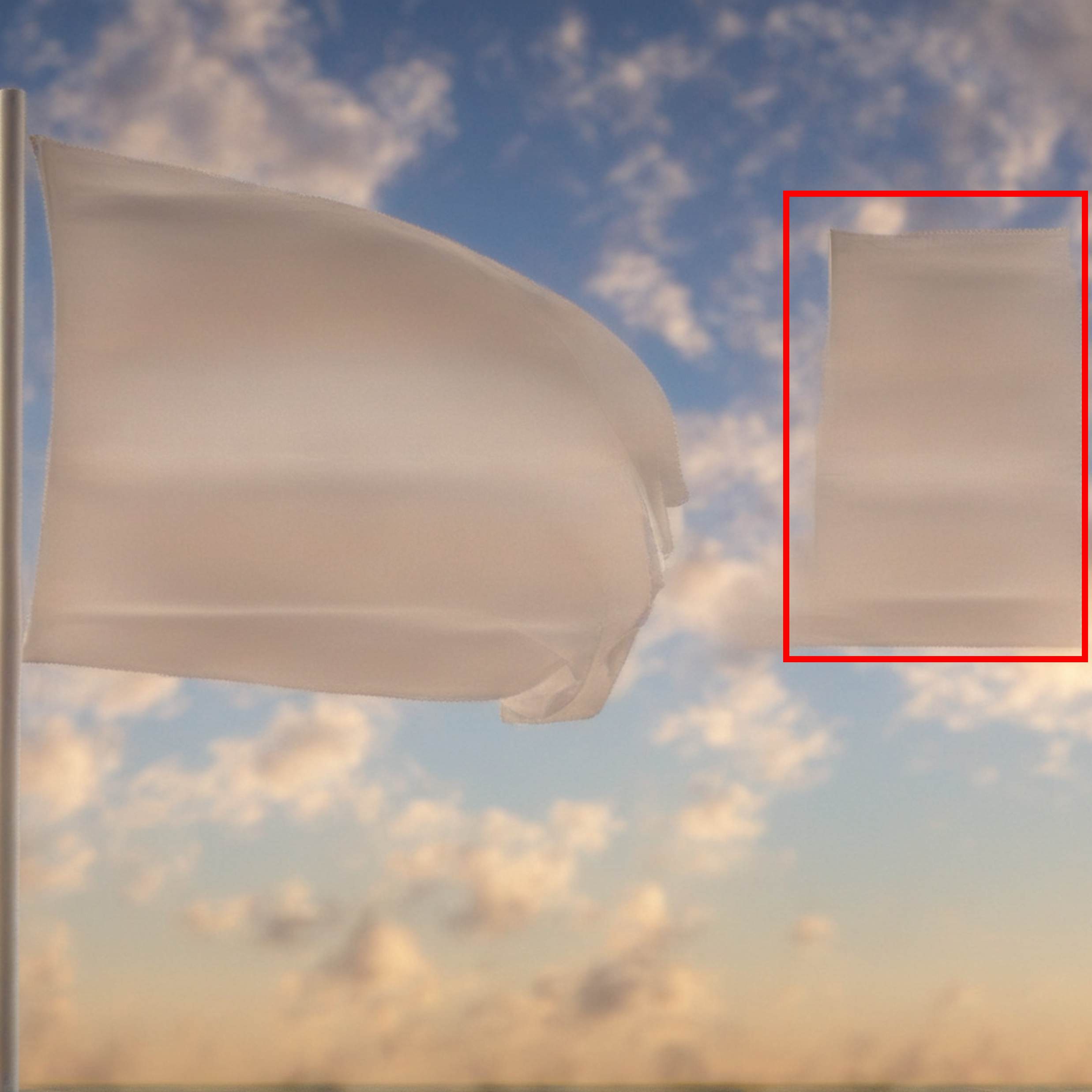}
\end{minipage}%
\begin{minipage}[t]{\minipagesize\textwidth}
  \centering
  \includegraphics[width=\imagesize\linewidth]{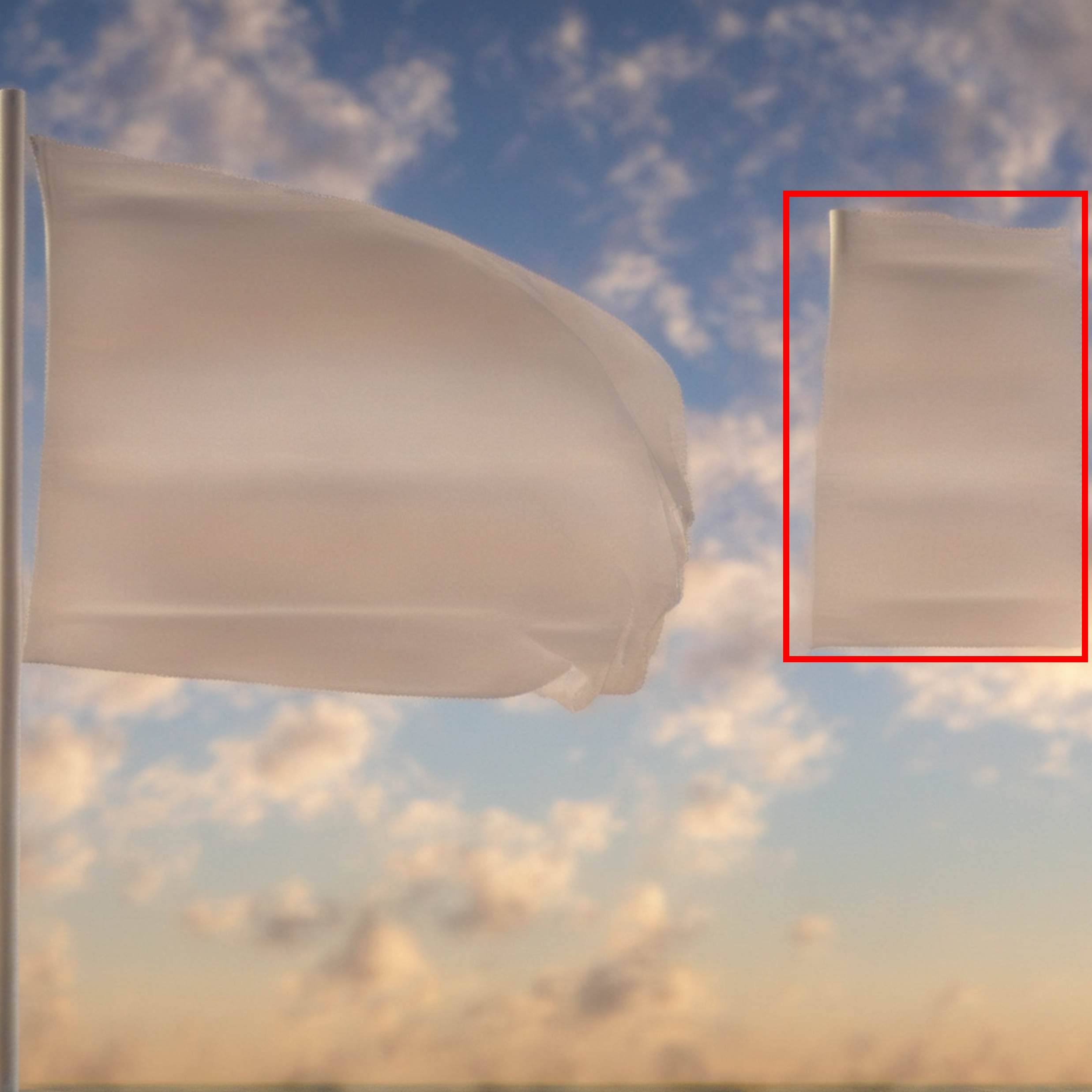}
\end{minipage}%
\hfill
\vspace{\vspacehill}
\hfill
\begin{minipage}[t]{\rotateboxsize\textwidth}
\rotatebox{90}{~~~~$\alpha = 0.75$}
\end{minipage}%
\begin{minipage}[t]{\minipagesize\textwidth}
  \centering
  \includegraphics[width=\imagesize\linewidth]{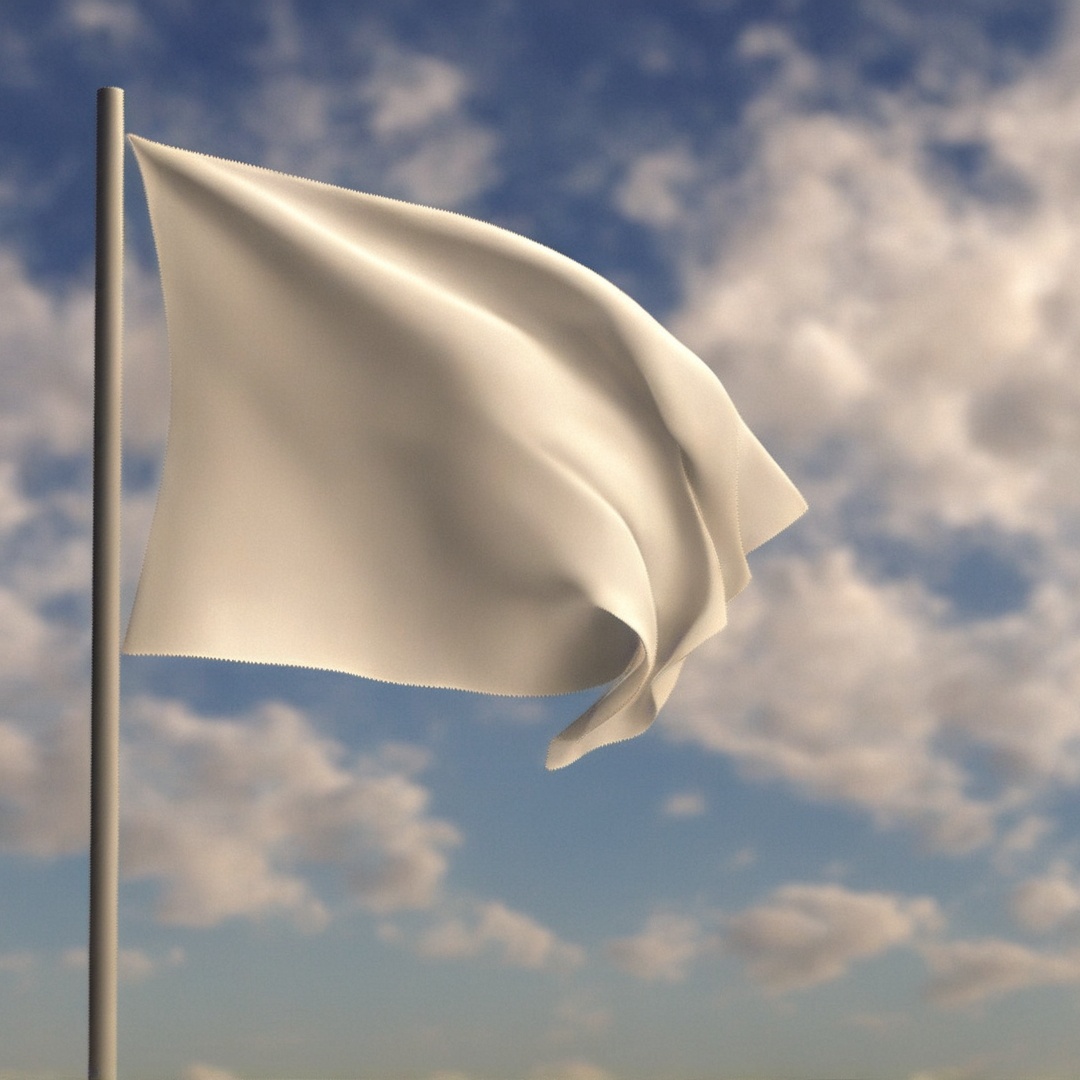}
\end{minipage}%
\begin{minipage}[t]{\minipagesize\textwidth}
  \centering
  \includegraphics[width=\imagesize\linewidth]{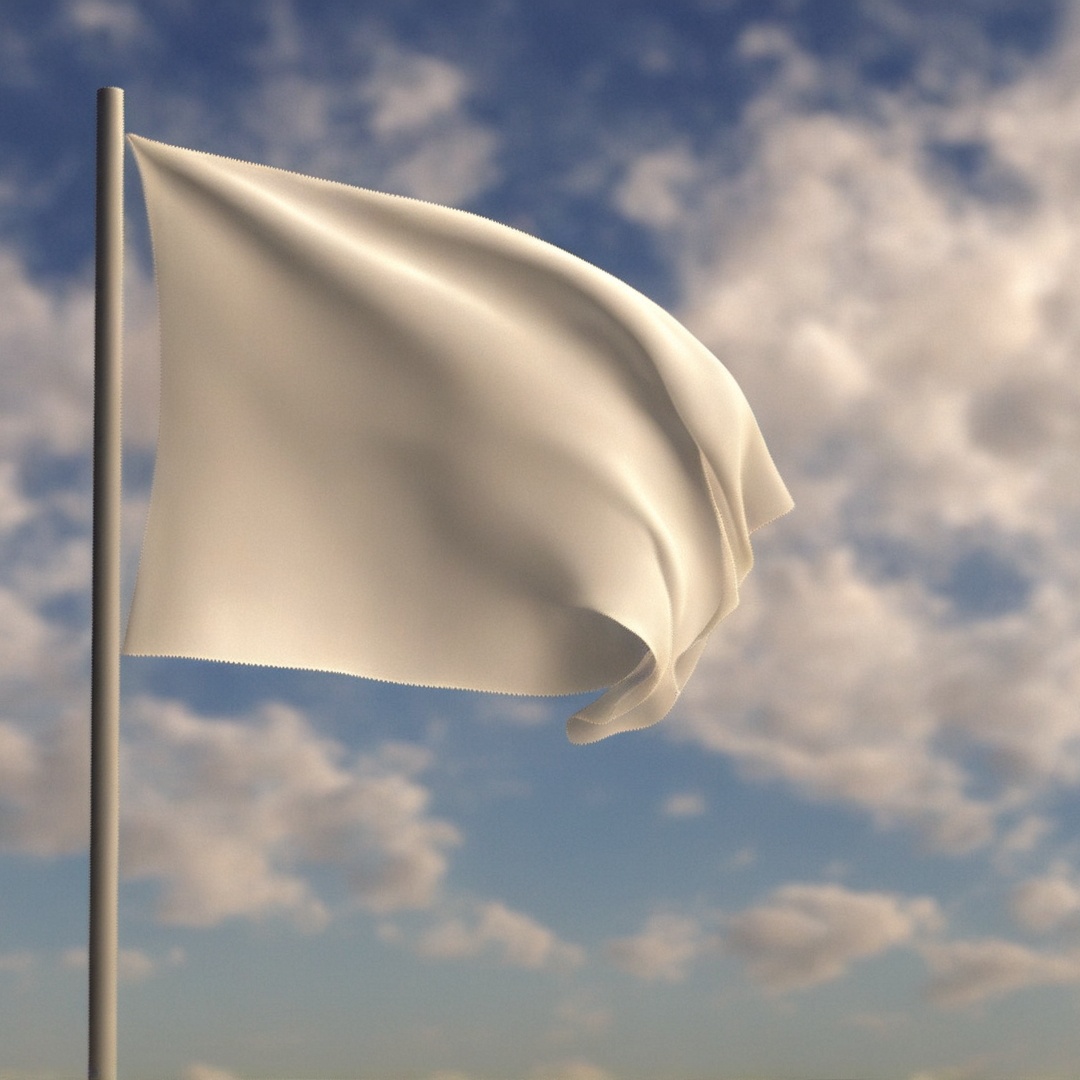}
\end{minipage}%
\begin{minipage}[t]{\minipagesize\textwidth}
  \centering
  \includegraphics[width=\imagesize\linewidth]{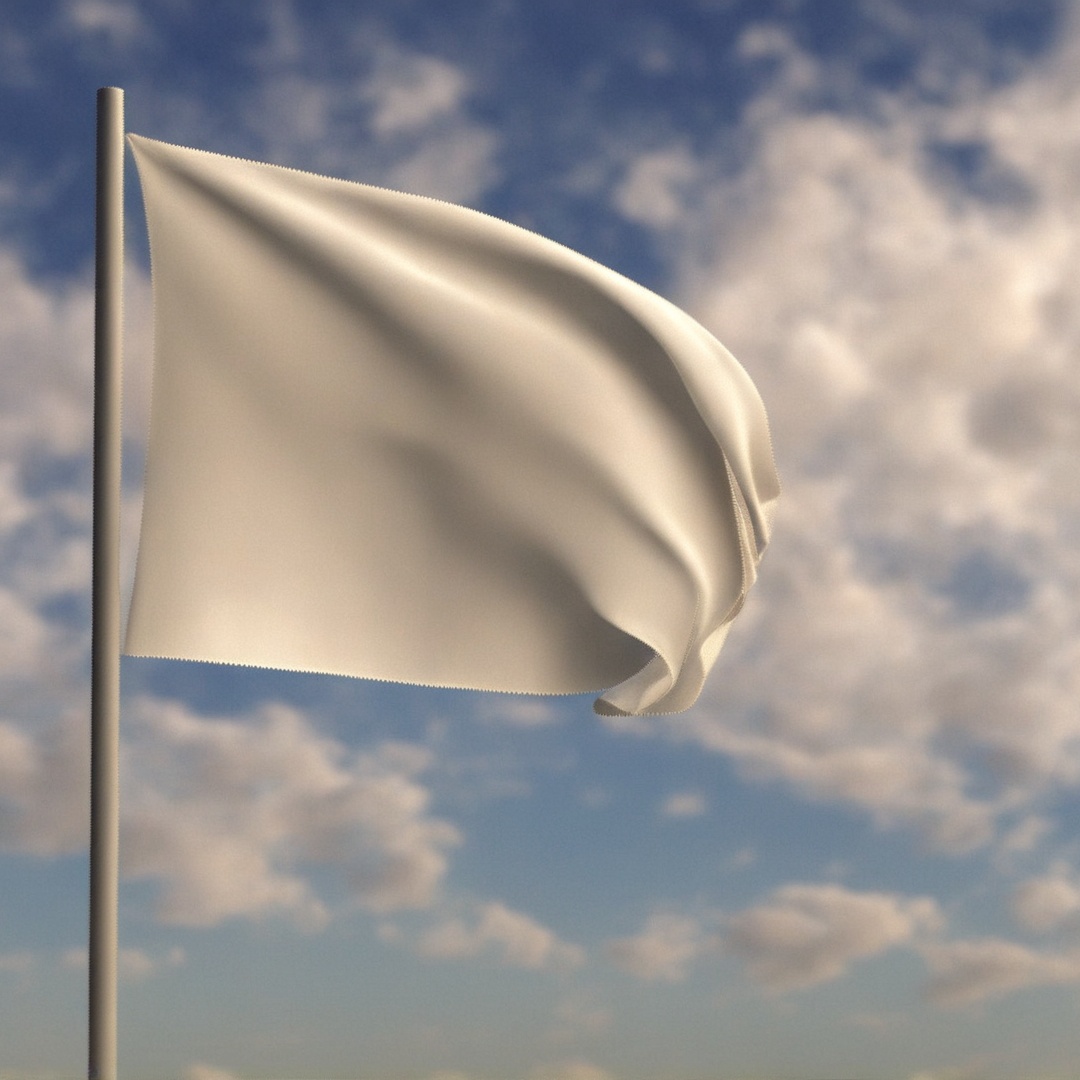}
\end{minipage}%
\begin{minipage}[t]{\minipagesize\textwidth}
  \centering
  \includegraphics[width=\imagesize\linewidth]{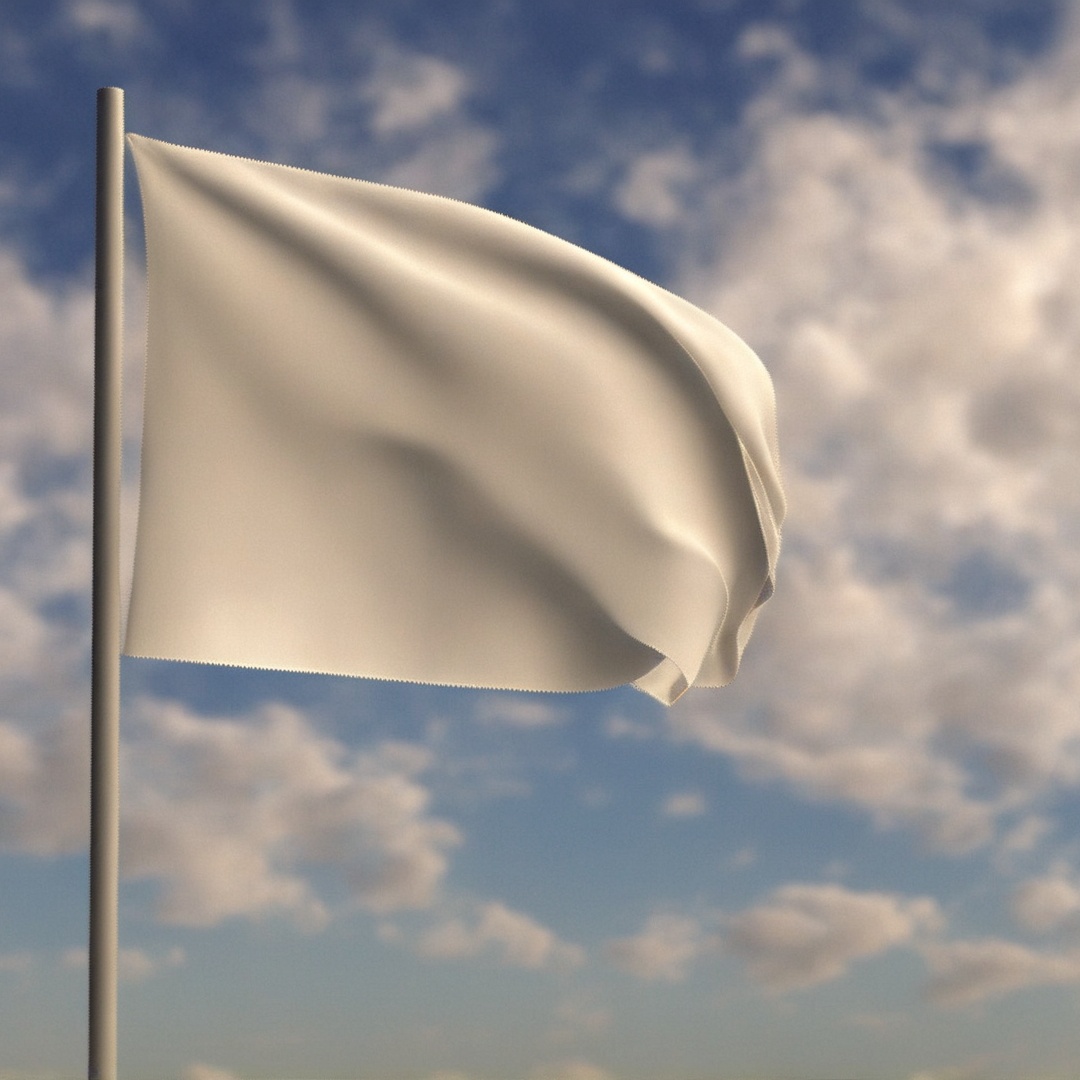}
\end{minipage}%
\hfill
\vspace{\vspacehill}
\hfill
\begin{minipage}[t]{\rotateboxsize\textwidth}
\rotatebox{90}{~~~~$\alpha = 1.0$}
\end{minipage}%
\begin{minipage}[t]{\minipagesize\textwidth}
  \centering
  \includegraphics[width=\imagesize\linewidth]{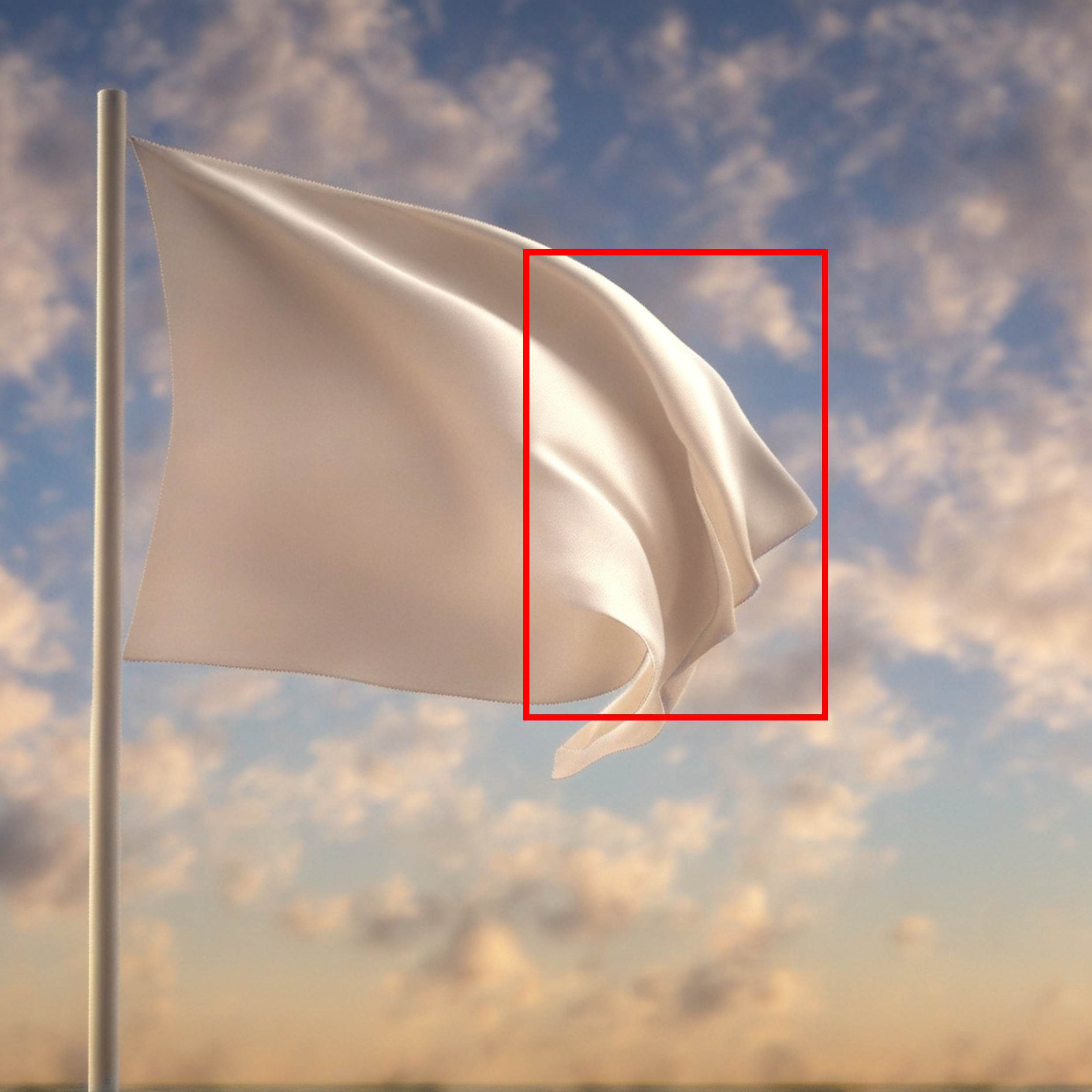}
\end{minipage}%
\begin{minipage}[t]{\minipagesize\textwidth}
  \centering
  \includegraphics[width=\imagesize\linewidth]{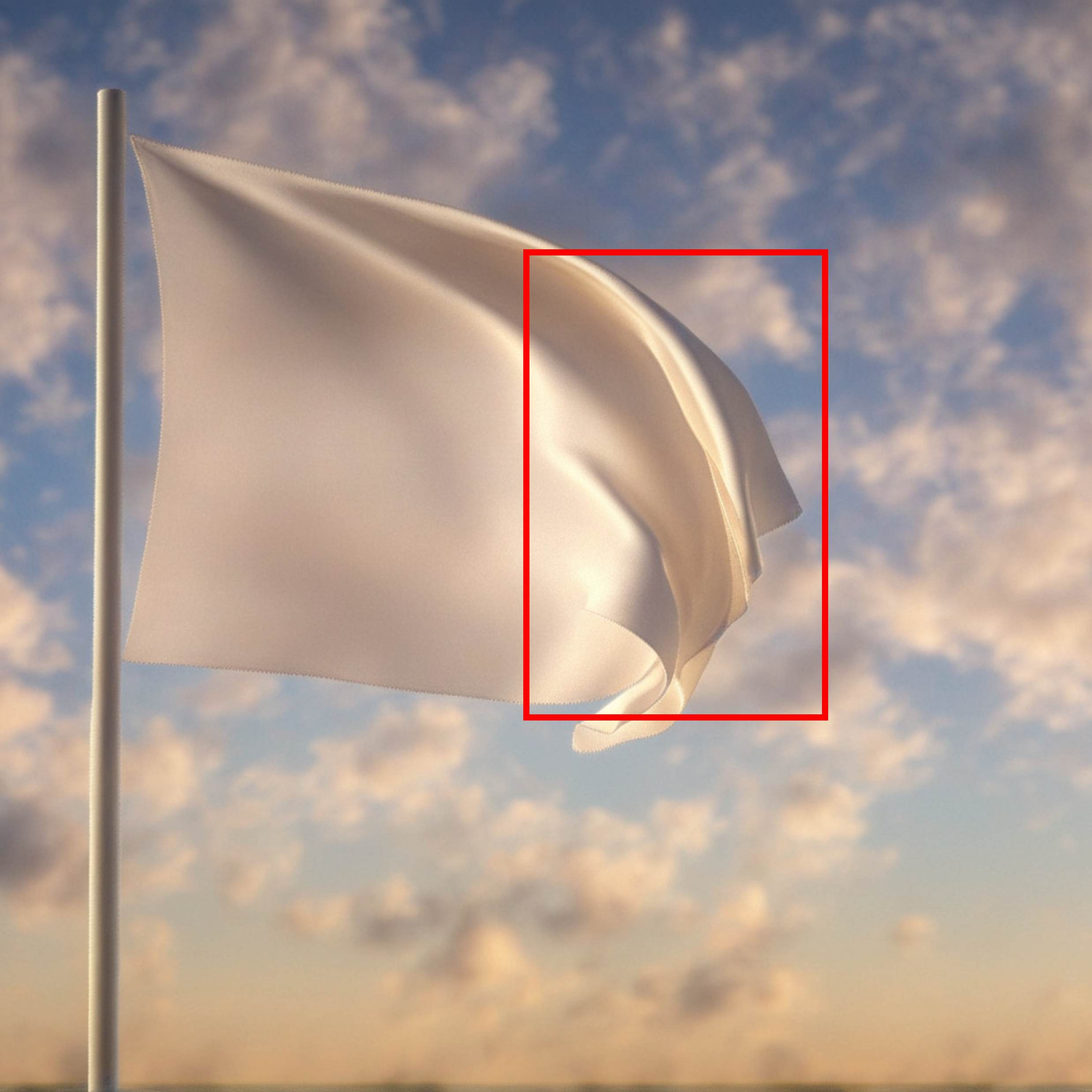}
\end{minipage}%
\begin{minipage}[t]{\minipagesize\textwidth}
  \centering
  \includegraphics[width=\imagesize\linewidth]{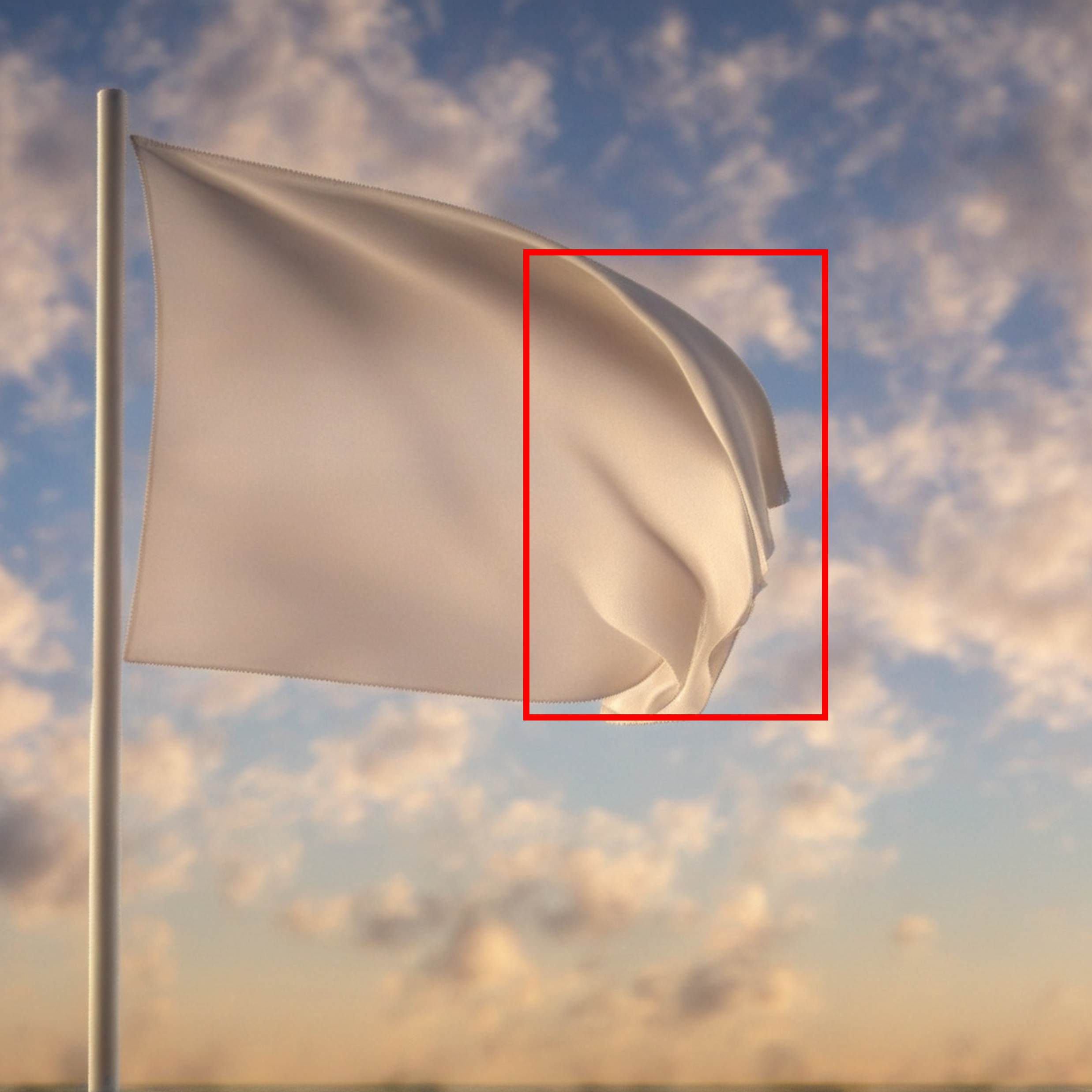}
\end{minipage}%
\begin{minipage}[t]{\minipagesize\textwidth}
  \centering
  \includegraphics[width=\imagesize\linewidth]{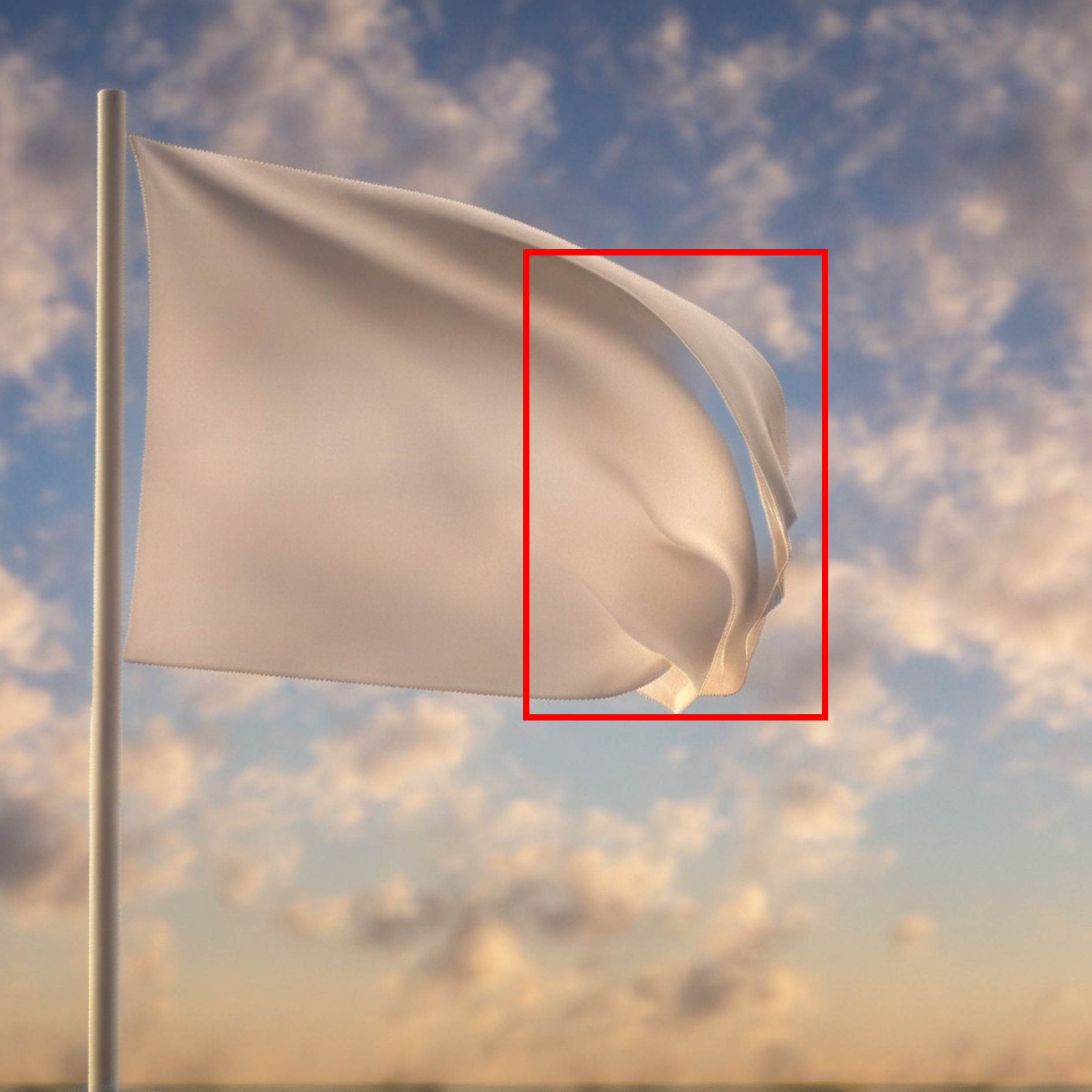}
\end{minipage}%
\hfill
\begin{minipage}[t]{\rotateboxsize\textwidth}
\centering
\rotatebox{90}{~}
\end{minipage}%
\begin{minipage}[t]{\minipagesize\textwidth}
  \centering
\small{$i$}
\end{minipage}%
\begin{minipage}[t]{\minipagesize\textwidth}
  \centering
\small{$i + 1$}
\end{minipage}%
\begin{minipage}[t]{\minipagesize\textwidth}
  \centering
\small{$i + 2$}
\end{minipage}%
\begin{minipage}[t]{\minipagesize\textwidth}
  \centering
\small{$i + 3$}
\end{minipage}%
\hfill
\vspace{-0.1cm}
    \caption{Ablation experiments on different $\alpha$ values. Four consecutive frames are shown.}
    \label{sec4:ablation_experiment_flag_2}
    \vspace{-0.1cm}
\end{figure}

\def\sizethree{0.15}
\def\maxzhen{100}
\def\jianxi{1mm}
\def\sizetwo{0.15}

\section{Limitations and Future Work}

Although \methodshort{} advances the field of T2V synthesis, it has several limitations that set the directions for future research. While \methodshort{} successfully handles basic physical motions related to specific object materials, we have not extended it to more complex motion scenarios. We hypothesize that complex motions could be decomposed into a series of basic motions, requiring more refined instructions for LLMs. 
Another limitation is that sometimes the generated videos still have flickering in some frames. Despite these limitations, we believe that \methodshort{} provides a promising way for T2V generation.

\section{Conclusions}

This paper proposes \methodshort{}, a new training-free framework that effectively combines the advanced planning capability of Large Language Models (LLMs) with the robust simulation tool, Blender, for efficient text-to-video (T2V) synthesis. By generating Blender's scripts via GPT-4, \methodshort{} significantly simplifies the video generation process, making it more accessible and less reliant on extensive manual effort or a deep, specialized technical knowledge in 3D modeling. Experimental results on three basic physical motion scenarios, including rigid object drop and collision, cloth draping and swinging, and liquid flow, demonstrate \methodshort{}'s impressive capability to efficiently generate high-quality videos with temporal coherence, surpassing previous T2V methods. \methodshort{} opens up new perspectives for T2V generation. Its integration of LLM-driven scripting and advanced Blender simulation paves a promising path for tackling more complex scenes in future research.

    \bibliographystyle{ieeenat_fullname}
    \bibliography{main}
\clearpage

\twocolumn[{%
	\renewcommand\twocolumn[1][]{#1}%
	
\def\sizefive{0.20}
\vspace{-5mm}
\setlength{\tabcolsep}{0.5pt}
\renewcommand{\arraystretch}{1.15}
\begin{tabular}{c c c c c}
	\multicolumn{5}{c}{\includegraphics[width=0.9\columnwidth]{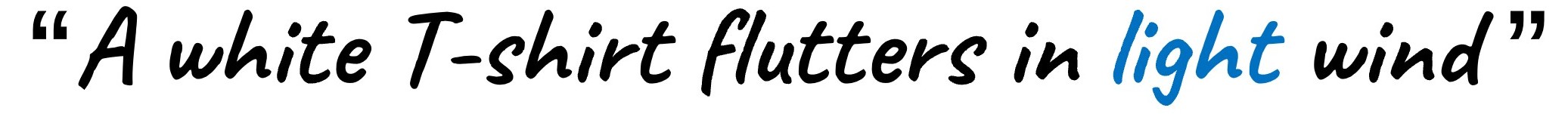}}\\
	\animategraphics[width=\sizefive\textwidth, autoplay, loop]{24}{video_imgs_supp/T-shirt-light-wind/GPT4Motion/}{0}{80} &
	\animategraphics[width=\sizefive\textwidth, autoplay, loop]{7}{video_imgs_supp/T-shirt-light-wind/animatediff/}{0}{15} &
	\animategraphics[width=\sizefive\textwidth, autoplay, loop]{7}{video_imgs_supp/T-shirt-light-wind/modelscope/}{0}{15} &
	\animategraphics[width=\sizefive\textwidth, autoplay, loop]{3}{video_imgs_supp/T-shirt-light-wind/t2v0/}{0}{7} &
	\animategraphics[width=\sizefive\textwidth, autoplay, loop]{3}{video_imgs_supp/T-shirt-light-wind/direcT2V/}{0}{7}\\
\end{tabular}
\vspace{-0.2cm}
\vspace{3mm}

\begin{tabular}{c c c c c}
	\multicolumn{5}{c}{\includegraphics[width=0.9\columnwidth]{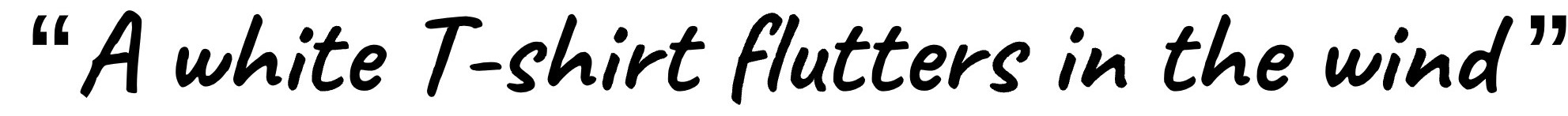}}\\
	\animategraphics[width=\sizefive\textwidth, autoplay, loop]{24}{video_imgs_supp/T-shirt-wind/GPT4Motion/}{0}{80} &
	\animategraphics[width=\sizefive\textwidth, autoplay, loop]{7}{video_imgs_supp/T-shirt-wind/animatediff/}{0}{15} &
	\animategraphics[width=\sizefive\textwidth, autoplay, loop]{7}{video_imgs_supp/T-shirt-wind/modelscope/}{0}{15} &
	\animategraphics[width=\sizefive\textwidth, autoplay, loop]{3}{video_imgs_supp/T-shirt-wind/t2v0/}{0}{7} &
	\animategraphics[width=\sizefive\textwidth, autoplay, loop]{3}{video_imgs_supp/T-shirt-wind/direcT2V/}{0}{7}\\
\end{tabular}
\vspace{-0.2cm}
\vspace{3mm}

\begin{tabular}{c c c c c}
	\multicolumn{5}{c}{\includegraphics[width=0.98\columnwidth]{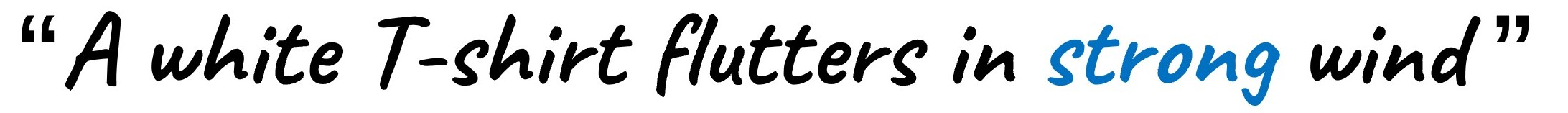}}\\
	\animategraphics[width=\sizefive\textwidth, autoplay, loop]{24}{video_imgs_supp/T-shirt-strong-wind/GPT4Motion/}{0}{80} &
	\animategraphics[width=\sizefive\textwidth, autoplay, loop]{7}{video_imgs_supp/T-shirt-strong-wind/animatediff/}{0}{15} &
	\animategraphics[width=\sizefive\textwidth, autoplay, loop]{7}{video_imgs_supp/T-shirt-strong-wind/modelscope/}{0}{15} &
	\animategraphics[width=\sizefive\textwidth, autoplay, loop]{3}{video_imgs_supp/T-shirt-strong-wind/t2v0/}{0}{7} &
	\animategraphics[width=\sizefive\textwidth, autoplay, loop]{3}{video_imgs_supp/T-shirt-strong-wind/direcT2V/}{0}{7}\\
	\multicolumn{1}{c}{\normalsize{GPT4Motion}}&\normalsize{AnimateDiff \cite{guo2023animatediff}}  &\normalsize{ModelScope \cite{wang2023modelscope}}  &\normalsize{Text2Video-Zero \cite{khachatryan2023text2video}} &\normalsize{DirecT2V \cite{hong2023large}}   \\
\end{tabular}
\vspace{-0.2cm}
\captionof{figure}{Comparison of the video results generated by different text-to-video models under different physical conditions. {\emph{Best viewed with \href{https://www.adobe.com/acrobat/pdf-reader.html}{Acrobat Reader} for animation. 
}}}
\vspace{3mm}
\label{suppl:comparison_experiment_T-shirt}

}]
\section{Supplementary Material}
In this supplement material, we further compare the generation ability of \methodshort{} and four baselines for different physical conditions in Sec.~\ref{supp:experiments}. In Sec.~\ref{supp:metrics}, we introduce the quantitative metrics we used. In Sec.~\ref{supp:AdvancedSettingsinBlender}, we describe the details of our settings for Blender. In Sec.~\ref{supp:GPT4_Encapsulated} and Sec.~\ref{supp:physicsknowledge}, we show how GPT-4 generates encapsulated Python functions via prompts, and how GPT-4 incorporates its own knowledge of physics to control the motion of objects, respectively. 

\subsection{More Comparison with Baselines}
\label{supp:experiments}
In the main paper, we have compared \methodshort{} with four baselines (AnimateDiff \cite{guo2023animatediff}, ModelScope \cite{wang2023modelscope}, Text2Video-Zero \cite{khachatryan2023text2video}, and DirecT2V \cite{hong2023large}) on three scenarios (rigid object drop and collision, cloth draping and swinging, and liquid flow). 
Here, we further conduct an experiment on dynamic effects of a T-shirt being blown by the wind under three wind strengths. 
The results are shown in Figure \ref{suppl:comparison_experiment_T-shirt}, where the seed is randomly chosen and fixed in all the generations. We can see that these baselines all fail to generate videos that match the user prompts and are unable to control the intensity of physical phenomena solely based on the linguistic descriptions. 
In contrast, our \methodshort{} not only precisely designs the parameters of Blender encapsulated functions (such as wind strength) through GPT-4, but also leverages Blender’s physics engine to simulate the complex flapping and twisting dynamics of the T-shirt in the wind.

\subsection{Quantitative Evaluation Metrics}
\label{supp:metrics}
Here, we introduce the metrics employed in the main paper:

\begin{enumerate}
    \item \textit{Motion Smoothness} \cite{huang2023vbench}. This metric evaluates the smoothness of motion in generated videos, ensuring it conforms to the physical laws of the real world. The evaluation utilizes motion priors from the video frame interpolation model \cite{li2023amt} to assess the smoothness of generated motions.
    
    \item \textit{Temporal Flickering} \cite{huang2023vbench}. This metric identifies imperfections in temporal consistency within generated videos, especially in local and high-frequency details. The method involves analyzing static frames and computing the mean absolute difference across frames. These values are then subtracted from 255 and normalized between 0 and 1.
    
    \item \textit{CLIP-Score} \cite{liu2023evalcrafter}. This metric is designed to assess Text-Video Consistency. It leverages a pretrained ViT-B/32 CLIP model \cite{radford2021learning} as a feature extractor to calculate the cosine similarity between each video frame and the associated text.
\end{enumerate}

\subsection{Blender Settings}
\label{supp:AdvancedSettingsinBlender}
We use Blender to generate two sequences of scene depth maps and edge maps. The edge maps are obtained by Blender's built-in Freestyle\footnote{\url{https://docs.blender.org/manual/en/latest/render/freestyle/introduction.html}} feature, which is an advanced rendering technique for non-photorealistic line drawings from 3D scenes. It not only allows for various line styles, such as hand-drawn or sketch, but also serves as an independent rendering pass without losing edge information of the scene and additional post-processing. Moreover, we normalize the scene's Z-channel and render it as an independent depth map. Since the rendering of edge maps and depth maps is unrelated to ray tracing, we choose Blender's ``Workbench" rendering mode. This mode avoids the substantial computational resources required for ray tracing, significantly enhancing rendering efficiency. In our experiments, completing the entire rendering process using Blender scripts takes only 2 to 3 minutes, with most of the time spent on calculating physical effects. 

\subsection{GPT-4's Generation of Encapsulated Python Functions for Blender}
\label{supp:GPT4_Encapsulated}
Blender incorporates an independent Python library, enabling the transformation of various manual processes within the software to automate tasks through its Python API. Leveraging GPT-4's capabilities of grasping Blender's Python API, we can rapidly generate encapsulated Python functions for Blender. This involves inputting specific functional goals and the associated Blender operation steps into GPT-4. Consequently, GPT-4 can autonomously create Python code that corresponds to these steps, effectively encapsulating them into Python functions. One example is given below.

\begin{quote}
\hspace{-2.3em}{\raisebox{-1.5ex}{\includegraphics[height=2em]{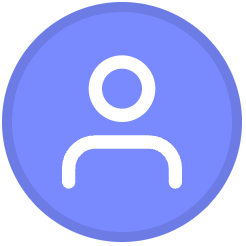}} \textbf{User:}} \\
\texttt{You are an expert in Blender and Python. Please generate encapsulated Blender Python functions by following our step-by-step instructions, writing the Blender Python API corresponding to each step, designing each function's name, and explaining its functionality and the meaning of each parameter in the docstring.}

\hspace{-2.3em}{\raisebox{-1.5ex}{\includegraphics[height=2em]{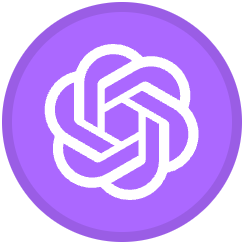}} \textbf{GPT-4:}} \\
\texttt{Sure, I'd be happy to help with that. Please go ahead and provide me with the step-by-step instructions for the Blender Python functions you need, and I will write the corresponding code for you.}

\hspace{-2.3em}{\raisebox{-1.5ex}{\includegraphics[height=2em]{User.png}} \textbf{User:}} \\
\texttt{Please build a function to remove all objects, cameras and lights from the initial scene.} 

\hspace{-2.3em}{\raisebox{-1.5ex}{\includegraphics[height=2em]{GPT.png}} \textbf{GPT-4:}} \\
\texttt{Certainly! To create a Blender Python function that removes all objects, cameras, and lights from the initial scene, you can define a function like this:}

\begin{lstlisting}
# bpy is the Python module used for scripting within Blender
import bpy

def clear_scene():
    """
    Clears all objects from the current Blender scene.
    
    This function selects all objects in the scene and deletes them.
    It is useful when starting a new scene setup or resetting the scene to a blank state.
    
    No inputs or outputs.
    
    Typically called at the beginning of a script when starting a new scene setup.
    """
    bpy.ops.object.select_all(action='SELECT')
    bpy.ops.object.delete()
\end{lstlisting}

\texttt{You can call this function whenever you need to clear the scene of all types of objects. Just make sure to be in the correct context where the scene is accessible.}

\hspace{-2.3em}{\raisebox{-1.5ex}{\includegraphics[height=2em]{User.png}} \textbf{User:}} \\
\texttt{Please now create two physics functions, one that adds collision properties to the inputted object, and another that adds rigid body properties to the object. Then, construct a function to create a floor that scales it up 50 times in all directions, with physics set to collision and rigid body (passive, with elasticity set to 1).}

\hspace{-2.3em}{\raisebox{-1.5ex}{\includegraphics[height=2em]{GPT.png}} \textbf{GPT-4:}} \\

\texttt{To accomplish this task, we'll create three functions in Blender's Python API:
\begin{enumerate}
    \item Add Collision Properties: This function will add collision properties to a specified object.
    \item Add Rigid Body Properties: This function will add rigid body properties to a specified object.
    \item Create a Floor with Physics: This function will create a floor, scale it, and set its physics properties to collision and rigid body.
\end{enumerate}
}

\begin{lstlisting}
import bpy

def add_collision(obj):
    """
    Adds a collision modifier to a Blender object if it does not already have one.

    Parameters:
    - obj (Blender Object): The object to which the collision modifier will be added.

    Returns:
    - None, but the object will have a collision modifier added to it if it wasn't present before.

    This function is used in physics simulations where it's necessary for objects to interact with each other, 
    such as rigid body or soft body simulations. The collision modifier makes the object a collider in the physics 
    simulation, allowing other objects to bounce off or slide along its surface.
    """
    if "Collision" not in obj.modifiers:
        bpy.ops.object.modifier_add(type='COLLISION')

def add_rigid_body(obj, mass=1, elasticity=0.5, rigid_body_type='ACTIVE'):
    """
    Adds a rigid body physics characteristic to a Blender object.
    
    Args:
    - obj (Blender Object): The object to which the rigid body physics will be applied.
    - mass (float): The mass of the object, default is 1.
    - elasticity (float): The restitution (bounciness) of the object, default is 0.5.
    - rigid_body_type (str): The type of rigid body, can be 'ACTIVE' or 'PASSIVE', default is 'ACTIVE'.
    
    Returns:
    - None, but the object is now a rigid body with the specified physics properties.
    
    This function is crucial for physics simulations, defining how the object behaves under physical forces.
    """
    bpy.ops.object.select_all(action='DESELECT')
    obj.select_set(True)
    bpy.context.view_layer.objects.active = obj
    bpy.ops.rigidbody.object_add()
    obj.rigid_body.type = rigid_body_type
    obj.rigid_body.mass = mass
    obj.rigid_body.restitution = elasticity
    obj.rigid_body.collision_shape = 'MESH'

def create_floor(elasticity=1):
    """
    Creates a floor plane in Blender, scales it, and sets it up with collision and rigid body physics.
    The created floor is scaled to be large enough to act as a ground plane for most scenes.

    Parameters:
    - elasticity (float): The restitution or 'bounciness' of the floor. A value of 1 means perfectly elastic, while 0 means no elasticity. Default is 1.
    """
    bpy.ops.mesh.primitive_plane_add(size=1, enter_editmode=False, align='WORLD', location=(0, 0, 0))
    floor = bpy.context.active_object
    floor.scale = (50, 50, 50)
    floor.name = 'GROUND'
    add_collision(floor)
    add_rigid_body(floor, rigid_body_type='PASSIVE', elasticity=elasticity)
\end{lstlisting}

\texttt{These functions together allow you to dynamically add physics properties to objects and create a large, physical floor in your Blender scene. Remember to run these functions in a Blender Python scripting context where the bpy module is accessible.}

\end{quote}

After generating the function, we enhance its usability for GPT-4 by incorporating instructions (such as the strength interval of the wind) in the function's docstring. This ensures that GPT-4's usage aligns more closely with our expectations.

\subsection{Combination of Blender and GPT-4's Physics Knowledge}
\label{supp:physicsknowledge}
We discover that \methodshort{} can not only use GPT-4's semantic understanding and coding capabilities to generate Blender Python scripts, but it can also leverage GPT-4's physical knowledge to control object motion. For example, when prompted with ``a basketball thrown towards the camera.", we find that GPT-4 can calculate the time it takes for the basketball to fall, based on the height difference between the initial position of the basketball and the camera. This calculation helps to determine the initial velocity required for the basketball to fly to the camera. The speed calculation code generated by GPT-4 is as follows:
\begin{lstlisting}
# Setup the basketball
basketball = setup_object(
    os.path.join(ASSETS_PATH, "basketball.obj"),
    "Basketball",
    size=0.24,  # Standard basketball diameter in meters
    mass=0.625,  # Standard basketball mass in kilograms
    position=(0, 0, 4)  # Initial position for the basketball
)

# Calculate the initial velocity for the basketball to hit the camera
g = 9.81  # Acceleration due to gravity in m/s^2
camera_height = 1.8521
height_difference = 4 - camera_height
time_to_fall = math.sqrt(2 * height_difference / g)
initial_horizontal_velocity = 13.665 / time_to_fall  # Distance to camera along Y-axis

# Set the initial velocity and rotation for the basketball
add_initial_velocity_for_rigid_body(
    basketball,
    initial_velocity=(0, initial_horizontal_velocity, 0),
    initial_rotation=(0, 0, 0)  # No initial rotation
)
\end{lstlisting}

In the above script, GPT-4 first calculates the height difference between the initial position of the basketball and the camera, which determines the time required for the basketball to fall to the same height as the camera. Subsequently, GPT-4 calculates the distance between the basketball and the camera along the Y-axis to determine the required initial velocity of the basketball. This process effectively integrates basic principles of physics, such as the equations of motion, to solve a practical problem in a simulated environment like Blender.

\end{document}